\documentclass{article}

\usepackage[preprint,nonatbib]{neurips_2025}

\usepackage{floatrow}
\newfloatcommand{capbtabbox}{table}[][\FBwidth]
\newfloatcommand{capbfigbox}{figure}[][\FBwidth]

\usepackage{enumitem}
\usepackage{pifont}
\usepackage{balance}
\usepackage{multicol}
\usepackage{multirow}
\usepackage{xcolor}
\usepackage{colortbl}
\definecolor{mygray}{gray}{.9}
\usepackage{caption}
\usepackage{makecell}
\usepackage{algorithm}
\usepackage{algorithmic}
\usepackage{graphicx}
\usepackage[utf8]{inputenc} % allow utf-8 input
\usepackage[T1]{fontenc}    % use 8-bit T1 fonts
\definecolor{cvprblue}{rgb}{0.21,0.49,0.74}
\usepackage[pagebackref,breaklinks,colorlinks,allcolors=cvprblue]{hyperref}
\usepackage{url}            % simple URL typesetting
\usepackage{booktabs}       % professional-quality tables
\usepackage{amsfonts}       % blackboard math symbols
\usepackage{nicefrac}       % compact symbols for 1/2, etc.
\usepackage{microtype}      % microtypography
\usepackage{xcolor}         % colors
\usepackage{titletoc}
\usepackage{amsmath}        % begin cases
\usepackage{wrapfig}        % begin cases

\usepackage[capitalize]{cleveref}

\usepackage[utf8]{inputenc} % allow utf-8 input
\usepackage[T1]{fontenc}    % use 8-bit T1 fonts
\usepackage{hyperref}       % hyperlinks
\usepackage{url}            % simple URL typesetting
\usepackage{booktabs}       % professional-quality tables
\usepackage{amsfonts}       % blackboard math symbols
\usepackage{nicefrac}       % compact symbols for 1/2, etc.
\usepackage{microtype}      % microtypography
\usepackage{xcolor}         % colors

\title{Time-R1: Post-Training Large Vision Language Model for Temporal Video Grounding}
\vspace{-0.6cm}
\author{
\vspace{-0.4cm}
\\ 
    Ye Wang$^{1*}$\quad Ziheng Wang$^{1*}$\quad Boshen Xu$^{1*\ddagger}$\quad Yang Du$^{1}$\quad Kejun Lin$^{1}$\quad Zihan Xiao$^{3}$\\ 
    Zihao Yue$^{1}$\quad Jianzhong Ju$^{2}$\quad Liang Zhang$^{1}$\quad Dingyi Yang$^{1}$\quad Xiangnan Fang$^{1}$\quad Zewen He$^{2}$\\
    Zhenbo Luo$^{2}$\quad Wenxuan Wang$^{1}$\quad Junqi Lin$^{2}$\quad Jian Luan$^{2}$\quad Qin Jin$^{1\dagger}$\\
    \\ 
    $^{1}$AIM3 Lab, Renmin University of China\\
    $^{2}$MiLM Plus, Xiaomi Inc. \quad $^{3}$Independent Researcher
    \\ 
    Project Page: \url{https://xuboshen.github.io/Time-R1/}
}

\begin{document}

\renewcommand{\thefootnote}%
{\fnsymbol{footnote}}
\footnotetext[0]{$\dagger$ Corresponding author: Qin Jin; $^*$ Equal contribution, listed in alphabetical order; $\ddagger$ Project lead.}

\maketitle
\begin{abstract}
Temporal Video Grounding (TVG), the task of locating specific video segments based on language queries, is a core challenge in long-form video understanding. While recent Large Vision-Language Models (LVLMs) have shown early promise in tackling TVG through supervised fine-tuning (SFT), their abilities to generalize remain limited. 
To address this, we propose a novel post-training framework that enhances the generalization capabilities of LVLMs via reinforcement learning (RL).
Specifically, our contributions span three key directions:
(1) \textbf{Time-R1}: we introduce a reasoning-guided post-training framework via RL with verifiable reward to enhance the capabilities of LVLMs on the TVG task. 
(2) \textbf{TimeRFT}: we explore data-efficient post-training strategies on our curated RL-friendly dataset, which trains the model to progressively comprehend difficult samples, leading to better generalization.
(3) \textbf{TVGBench}: we carefully construct a small yet comprehensive benchmark for LVLM evaluation, assessing 11 types of queries and featuring balanced distributions across both videos and queries.
Extensive experiments demonstrate that Time-R1 achieves state-of-the-art performance across multiple downstream datasets using only 2.5K training data, while improving its general video understanding capabilities. 

\end{abstract}

\begin{figure}[h]
  \centering
  \includegraphics[width=0.95\linewidth]{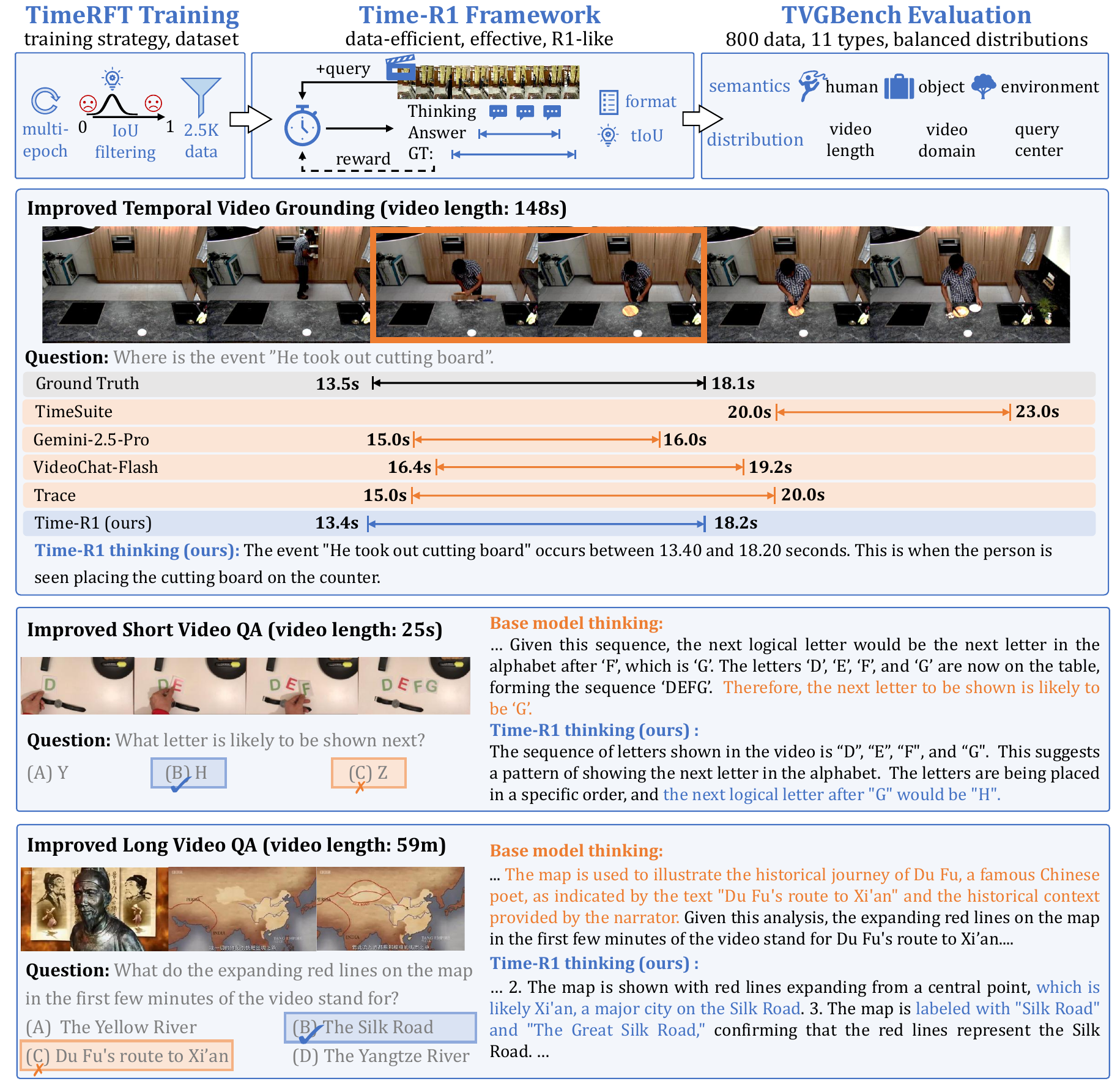}
  \caption{
  Our contributions include a novel post-training framework for LVLMs via reinforcement learning, \textbf{Time-R1}; an RL fine-tuning strategy along with its associated training dataset, \textbf{TimeRFT}; and a new benchmark, \textbf{TVGBench}, for evaluating LVLMs on the TVG task. 
  Our Time-R1 model not only achieves SoTA TVG performance, but also enhances performance on both short- and long-form multi-choice video question answering tasks.
  }
  \label{fig:placeholder}
\end{figure}

% Introduction
\section{Introduction}
\label{sec:intro}
Understanding long-form videos has long been a core ambition in computer vision~\cite{gaidon2013temporal,laptev2007retrieving,darrell1993space}. 
A critical step toward this goal is Temporal Video Grounding (TVG)~~\cite{gao2017tall,zhang2023tsgsurvey}, which enables models to localize video segments corresponding to natural language queries (e.g., ``Find the segment where a person walks into the living room."). 
Such capability is fundamental for real-world applications, including smart home assistants~\cite{yang2025egolife,grauman2022ego4d,sigurdsson2016Charades} and video retrieval systems on online platforms~\cite{caba2015activitynet,anne2017didemo}.

Traditional TVG approaches adopt a feature-based paradigm, where pretrained models (e.g., CLIP~\cite{radford2021clip}, I3D~\cite{Carreira_2017_CVPR_I3D}) extract text and video features, followed by task-specific grounding models~\cite{lin2023univtg,jang2023knowing,lin2022egovlp}. 
However, these methods suffer from error accumulation due to imperfect pretrained features.
To overcome these limitations, recent efforts have shifted toward end-to-end Large Vision-Language Models (LVLMs)~\cite{ren2024timechat,zeng2025timesuite}, which directly process long-form videos and text queries. 
Despite being pretrained on datasets 100× larger than domain-specific benchmarks~\cite{sigurdsson2016Charades}, LVLMs (with 7B+ parameters) often underperform compared to much smaller feature-based models (e.g., 9M parameters~\cite{jang2023knowing}). This raises a critical question: Why do LVLMs, despite their vast pretrained knowledge, fail to excel on TVG? 

We attribute the suboptimal performance of LVLMs to over-penalization of false negatives during supervised fine-tuning (SFT). 
For instance, when the ground truth timestamp is [2s, 4s], even when the model makes a reasonable prediction of timestamp [1.9s, 3.9s], the autoregressive loss would still be undesirably high. 
Such disproportionate penalties on reasonable predictions result in overfitting and poor generalization. 
While previous solutions have attempted to address this by introducing new timestamp tokens into the vocabulary~\cite{guo2024trace, guo2025vtgllm, yang2023vid2seq} or by appending a regression head to predict timestamps~\cite{zhao2025videoexpert}, they often sacrifice the pretrained language capabilities of LLMs.

Inspired by recent success in reinforcement learning (RL) for post-training LLMs~\cite{openaio1, deepseekr1} with chain-of-thought (CoT) prompting, especially in domains with deterministic answers, such as code generation and mathematical reasoning, we explore whether RL can serve as an effective alternative for TVG. Unlike SFT, RL allows direct optimization of task-specific metrics (e.g., IoU), thereby reducing rigid penalties of autoregressive losses and encouraging plausible timestamp predictions. 
In this work, we present an RL-based framework, Time-R1, that effectively post-trains LVLMs for TVG and pushes the performance frontier. Our contributions include:

\begin{itemize}[nosep,wide,labelindent=4pt,labelwidth=*,align=left]
\item\textbf{RL-based framework for temporal video grounding.} 
We introduce \textbf{Time-R1}, a reasoning-enhanced post-training framework via RL with verifiable rewards, where the LVLM first generates chain-of-thought descriptions and then predicts timestamps.
The post-training process is optimized using Generalized Reinforcement Policy Optimization (GRPO) with a novel reward function, incorporating both a structured template reward and a timestamp-aware tIoU reward.

\item\textbf{Time-aware reinforcement fine-tuning.} 
We propose \textbf{TimeRFT}, a reinforcement fine-tuning strategy with dynamic hard sampling, which mines hard samples on a curated dataset and progressively selects low-IoU samples for multi-epoch training.
To ensure stable reasoning and reduce hallucinations, we adopt a cold-start approach to generate CoT with video captions.
To support RL-friendly training, we curate an RFT dataset with difficulty annotations on the TVG task. % and multiple-choice video QA tasks.

\item\textbf{Comprehensive benchmark for LVLMs on TVG.} 
Existing TVG benchmarks are designed for the large-scale evaluation of small models.
Considering the inference speed bottlenecks and general-purpose role of LVLMs, we construct \textbf{TVGBench}, a compact yet comprehensive benchmark for TVG. 
We carefully balance the video distribution, query distribution, and design specific query semantics to ensure that the benchmark is well-suited for evaluating LVLMs.

\item\textbf{State-of-the-Art results and generalization.} 
Compared with 7B LVLMs on the temporal video grounding task, our method outperforms all prior SFT-based methods with only 2.5K training data. 
After fine-tuning on downstream benchmarks like Charades~\cite{sigurdsson2016Charades} and ActivityNet~\cite{caba2015activitynet}, it surpasses many previous feature-based approaches. 
Furthermore, Time-R1 also improves the general video understanding on video QA benchmarks like MVBench~\cite{li2024mvbench} and VideoMME~\cite{fu2024videomme}.
\end{itemize}
\begin{figure}[t]
  \centering
  \includegraphics[width=0.95\linewidth]{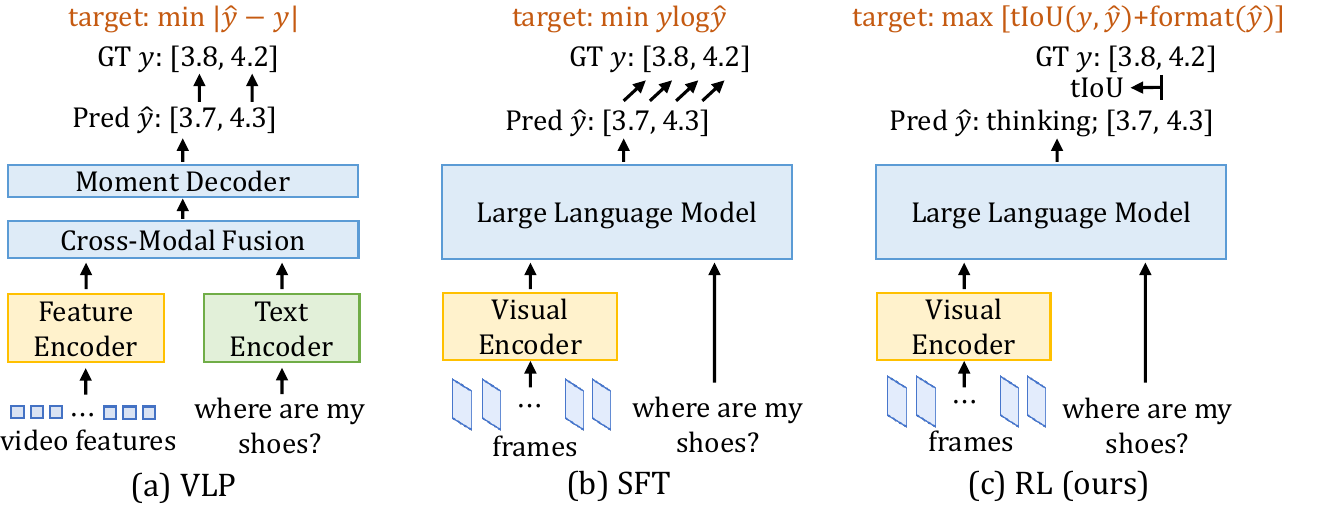}
  \vspace{-0.1cm}
   \caption{
   Comparison of different approaches for the TVG task, including feature-based video-language pretraining (VLP)~\cite{CVPR2024SnAG,jang2023knowing}, supervised fine-tuning (SFT)~\cite{qu2024chatvtg,zeng2025timesuite}, and RL (ours).}
   \label{fig:method_comparison}
   \vspace{-0.1cm}

\end{figure}

\section{Related Works}
\label{sec:related_works}

\noindent\textbf{Temporal video grounding.}
The TVG task~\cite{gao2017tall,anne2017didemo} aims to localize temporal segments in untrimmed long videos given natural language queries.
Previous works can be broadly categorized into feature-based video-language pretraining and frame-based LVLM methods, as shown in~\Cref{fig:method_comparison}.
Feature-based methods first extract video and language features using pre-trained encoders (e.g., I3D~\cite{Carreira_2017_CVPR_I3D}, EgoVLP~\cite{lin2022egovlp}, CLIP~\cite{radford2021clip}, BERT~\cite{devlin2019bert}, etc.), then build timestamp prediction modules based on multimodal fused features. Existing methods differ mainly in their design of the multimodal fusion module and timestamp prediction module. 
For example, SnAG~\cite{CVPR2024SnAG} adopts a late fusion strategy and regresses timestamps directly in a single forward pass without proposal generation. While effective, these approaches are fundamentally limited by the quality of the pretrained features.
Recent efforts have shifted toward end-to-end frame-based methods by fine-tuning LVLMs using SFT with autoregressive losses~\cite{li2024videochatflash,ren2024timechat,zeng2025timesuite,hannan2024revisionllm,wu2024numberit,guo2024trace,li2025imove}. For instance, TRACE~\cite{guo2024trace} treats each event as a combination of timestamp, saliency score, and caption, and fine-tunes the LVLM to generate event sequences autoregressively. However, such methods still underperform on even simple TVG benchmarks like Charades~\cite{sigurdsson2016Charades}, often lagging behind feature-based approaches.
In this work, we propose a novel RL-based post-training framework that establishes new state-of-the-art performance for LVLMs on TVG tasks, even surpassing many feature-based methods.

\noindent\textbf{RL in LLMs and LVLMs.}
RL is a foundational machine learning paradigm applied in diverse domains such as game playing~\cite{silver2017alphazero}, robotics~\cite{margolis2023walk}, and increasingly, language and vision-language models~\cite{openaio1}. RL updates models by interacting with environments and maximizing reward signals.
In recent years, RL has profoundly affected the field of LLM and LVLM post-training, which falls into two main categories: Reinforcement Learning with Human Feedback (RLHF)~\cite{ouyang2022rlhf,yu2024rlhfv} and Reinforcement Learning with Verifiable Reward (RLVR)~\cite{deepseekr1,openaio1,chen2025r1v}. 
% explanation of RLHF
Early works find that RLHF can align LLM to generate human preference data, which primarily reduces the safety risks in LLM and hallucination problems in LVLM. 
For example, RLHF-V~\cite{yu2024rlhfv} collects fine-grained pairs of incorrect and corrected captions and leverages direct preference optimization to optimize the model to generate correct captions, thus mitigating hallucinations.
% explanation of RLVR
More recent works have explored RLVR in tasks with deterministic answers, which not only benefits mathematical problem solving and code generation in LLMs (e.g., DeepSeek-R1~\cite{deepseekr1}), but also enhances the generalization of LVLMs across a range of applications, such as visual grounding~\cite{liu2025visualrft} and visual reasoning~\cite{tan2025reasonrft}.
As a pioneer among open-source LLMs, DeepSeek-R1~\cite{deepseekr1} adopts GRPO to enhance reasoning capabilities by designing rule-based rewards that incorporate both reasoning templates and final answers. In the context of LVLMs, MM-Eureka~\cite{meng2025mmeureka} investigates multimodal image reasoning using GRPO, introducing an online filtering mechanism and a two-stage training strategy to stabilize the optimization process.
However, existing approaches remain confined to language~\cite{deepseekr1,yu2025dapo}, image understanding~\cite{chen2025r1v,tan2025reasonrft,liu2025visualrft,meng2025mmeureka}, and short video understanding~\cite{zhang2025tinyllavavideor1,liao2025improved}. It remains unclear whether and how reinforcement learning impacts long-form video understanding. To bridge this gap, we take a first step by introducing RLVR into LVLMs for the temporal video grounding task.

\section{Method}
\label{sec:method}

The TVG task aims to temporally localize video segments within long-form videos based on natural language queries. 
Given a video of duration $t$ seconds, which is represented as a sequence of $T$ frames $\{x_1,\dots,x_T\}$, and a language query $q$, the goal is to identify the temporal boundaries $[t_s,t_e]$ of the segment that best corresponds to $q$, where $t_s, t_e\in \mathbb{R}^+$.
In this work, we introduce Time-R1, a framework designed to unleash the potential of LVLMs for the TVG task using RL.
We first provide background on RL-based training for LLMs in~\Cref{subsec: background}, then detail the training procedure of Time-R1 in~\Cref{subsec: Time-R1}.
Next, we describe specific training techniques used in TimeRFT in~\Cref{subsec: timerft}, and finally, we present the construction of our evaluation benchmark TVGBench in~\Cref{subsec: tvgbench}.

\subsection{Background of GRPO: RL for LLM}
\label{subsec: background}

As a pioneer among open-sourced R1-style LLMs, Deepseek-R1~\cite{deepseekr1} leverages GRPO to train the policy model $\pi_\theta$ (i.e., the LLM) to think before answering, making it particularly well-suited for tasks with well-defined answers, such as mathematical reasoning.
In the GRPO framework, given an input question $p$, the LLM samples $G$ candidate responses $o=\{o_1, \dots,o_G\}$, and a reward function $r(\cdot)$ assigns a reward score to each response, yielding $\{r(o_1), \dots, r(o_G)\}$.
GRPO encourages the LLM to generate responses that maximize a weighted sum reward $R(o)$, defined by:
\begin{equation}
\label{eq:ro}
    R(o)=\sum_{i=1}^G\frac{\pi_\theta (o_i)}{\pi_{\theta_{\mathrm{old}}}(o_i)} \cdot
    \frac{r(o_i)-\mathrm{mean}(\{r(o_i)\}_{i=1}^G)}{\mathrm{std}(\{r(o_i)\}_{i=1}^G)}
\end{equation}
where $\pi_\theta(o)$ denotes the probability of LLM generating the response $o$, and $\pi_{\theta_{\mathrm{old}}}$ represents the LLM parameters from a recently optimized state.
To ensure training stability and avoid large deviations from the original language model behavior, the final training objective incorporates a KL-divergence regularization term~\cite{deepseekr1}, penalizing divergence between $\pi_\theta$ and $\pi_\mathrm{ref}$:
\begin{equation}
\label{eq:grpo}
    \max_{\pi_\theta} \mathbb{E}_{o\sim \pi_{\theta_{\mathrm{old}}}(p)} [
        R(o) - 
        \beta \mathrm{D}_\mathrm{KL}(\pi_\theta \| \pi_\mathrm{ref})
    ] \\
\end{equation}
where $\beta$ is a scaling coefficient.
We omit the clipping operation for simplicity.

\subsection{Time-R1: RL for Temporal Video Grounding}
\label{subsec: Time-R1}
% Why RL
Since the TVG task has defined answers and well-established evaluation metrics, RL can optimize LVLMs for task-specific performance through tailored reward design.
To enhance interpretability and align with human-like reasoning, we additionally incorporate an explicit ``thinking process” prior to timestamp prediction. This process encourages the model to produce contextualized video descriptions that support its final decision.
We detail our reward modeling and training process below.

\noindent\textbf{Reward modeling.} 
The reward $r_i$ plays a crucial role in guiding the model’s learning objective.
To encourage effective temporal grounding with an explicit reasoning process, we design a composite reward function comprising two components: the timestamp-aware Intersection over Union (IoU) $r_{\mathrm{tIoU}}$ and the reasoning template reward $r_{\mathrm{form}}$.

\begin{itemize}[nosep,wide,labelindent=0pt,labelwidth=*,align=left]

    \item \textbf{Timestamp-aware IoU reward} $r_{\mathrm{tIoU}}(\cdot)$.
    The TVG task primarily uses IoU~\cite{yuan2021tIoU} to evaluate the quality of predicted segments against the ground-truth $[t_s', t_e']$, computed as:
    \begin{equation}
        \mathrm{IoU}=\frac{[t_s,t_e]\cap[t_s',t_e']}{[t_s,t_e]\cup[t_s',t_e']}
    \end{equation}
    where $A\cap B$ and $A\cup B$ denote the union and intersection between sets A and B, respectively.
    Optimizing for the IoU inherently encourages the LVLM to produce predictions that fall within a permissible range of variation $\epsilon$, such that $t_{s\ \mathrm{or}\ e}' - \epsilon \leq t_{s\ \mathrm{or}\ e} \leq t_{s\ \mathrm{or}\ e}' + \epsilon$ still yields high IoUs. 
    This encourages the LVLM to focus more on the semantic understanding of the event within possible temporal boundaries, rather than rigidly requiring exact temporal alignment like SFT.
    However, standard IoU may fail to accurately reflect the quality of temporal alignment in certain scenarios. 
    For example, when the ground truth span is [0, 30] (i.e., the full video), any prediction covering more than 30\% of the video would result in an IoU greater than 0.3. A prediction like [10, 25] would yield an IoU of 0.5, which overestimates its quality despite incorrect timestamps.    
    To address this issue, we introduce the timestamp-aware IoU (tIoU) as a corrective measure.
    tIoU augments the standard IoU with penalties on timestamp deviations, defined as:
    \begin{equation}
        r_{\mathrm{tIoU}}(o) = \mathrm{IoU}\cdot (1-\frac{|t_s-t_s'|}{t})\cdot (1-\frac{|t_e-t_e'|}{t})
    \end{equation}
    This modification penalizes predictions that deviate from the reference timestamps relative to the video duration $t$. In the earlier example, the reward value changes from 0.5 (IoU) to 0.28 (tIoU), providing a more realistic signal for learning.
    Overall, tIoU acts as a stricter and more informative reward signal, encouraging the LVLM to develop a deeper temporal understanding of events, rather than relying on superficial shortcuts.

    \item \textbf{Reasoning template reward} $r_{\mathrm{form}}(\cdot)$.
In TVG, the video segments relevant to a textual query typically comprise only a small portion of the entire long video. 
For LVLMs, it is therefore suboptimal to directly predict timestamps without first engaging in a reasoning process to identify the relevant content. Instead, the model should allocate its computational capacity toward reasoning over visual and linguistic cues to better understand the temporal context before making predictions. 
For instance, given the query ``the man washes dishes'', reasoning that the person is likely in a kitchen can improve temporal localization. Such context-aware inference supports more accurate and semantically aligned predictions.
To encourage this behavior, we introduce a template-based reasoning reward, which incentivizes the model to generate intermediate reasoning steps (structured in a predefined format) prior to timestamp localization.
The reasoning template reward requires the LVLM to structure its response like ``$<$think$>$$\cdots$$<$/think$>$ $<$answer$>$$<$$t_s$ to $t_e$$>$$<$/answer$>$'', formulated as:
    \begin{equation}
        r_{\mathrm{form}}(o) = 
            \begin{cases}
                 0, \mathrm{if}\ o\mathrm{\ has\ wrong\ fromat}   \\
                 1, \mathrm{if}\ o\mathrm{\ has\ correct\ fromat}
            \end{cases}
    \end{equation}

\end{itemize}

The overall reward is the sum of the two:
\begin{equation}
    r(o) = r_{\mathrm{tIoU}}(o) + r_{\mathrm{form}}(o)
\end{equation}

\noindent\textbf{GRPO training.}
The LVLM $\mathcal{F}(\cdot)$ takes the video frames ${x_1, \dots, x_t}$ and the language query $q$ as input and generates $G$ candidate responses ${o_1, \dots, o_G}$, where each response is computed as $o_i = \mathcal{F}(x_1, \dots, x_t; q)$. 
The reward for each response is calculated using Equation~\ref{eq:ro}, and the model is optimized with the GRPO objective in Equation~\ref{eq:grpo}.
To focus learning on the reasoning and localization capabilities, we freeze the visual encoder and update only the parameters of the LLM during training.

\subsection{TimeRFT: Time-Aware RL-Friendly Fine-Tuning}
\label{subsec: timerft}

Due to the high computational cost associated with RL training, we explore data-efficient strategies to reduce sample requirements.
To this end, we propose TimeRFT, which incorporates time-aware, RL-friendly dataset curation and fine-tuning techniques aimed at enhancing generalization while minimizing training overhead.

\noindent\textbf{RL-friendly dataset curation.}  We construct the TimeRFT dataset by leveraging only TVG samples and assign a difficulty score to each sample based on the base model's performance. A small subset is then selected for subsequent RL training.
\begin{itemize}[nosep,wide,labelindent=0pt,labelwidth=*,align=left]
    
    \item \textbf{Source data collection.}
    Our training videos are sourced from Internet video datasets including YT-Temporal~\cite{yang2023vid2seq_yt_temporal}, DiDeMo~\cite{anne2017didemo}, QuerYD~\cite{oncescu2021queryd}, InternVid~\cite{wang2023internvid}, and HowTo100M~\cite{miech2019howto100m}. 
    We obtain grounding data with annotations from VTG-IT~\cite{guo2025vtgllm}, TimeIT~\cite{ren2024timechat}, TimePro~\cite{zeng2025timesuite}, HTStep~\cite{afouras2023htstep}, and LongVid~\cite{li2024videochatflash}.
    This process yields 339K temporal grounding samples.
    
    \item \textbf{RFT data filtering.} 
    We propose a data selection strategy based on training difficulty to significantly reduce training costs while preserving strong generalization performance.
    Models trained only on easy samples (e.g., $\mathrm{IoU} \ge 0.7$) tend to overfit, whereas training on overly difficult samples (e.g., $\mathrm{IoU} = 0$) often suffers from sparse reward signals, making it hard for the model to receive positive feedback. To strike a balance, we select samples of moderate difficulty that are more conducive to generalization during reinforcement fine-tuning.
    We first estimate a difficulty score for each sample based on the performance of the base model. For grounding tasks, difficulty is quantified using the IoU between the predicted and ground-truth temporal regions.
    % For multi-choice QA, it is based on the model-assigned probability for the correct answer. 
    We then filter out samples that are either too easy or too hard. Specifically, we sample a subset of data from a Gaussian distribution over the IoU axis centered at 0.3, resulting in a set of 2.5K moderately difficult samples for RL training.

\end{itemize}

\noindent\textbf{RFT training strategy.} 
For selected difficult samples, the model may struggle to learn them in a single pass. However, we argue that effectively mastering these challenging cases is essential for improving overall model performance. To this end, we employ a multi-epoch training approach combined with per-epoch sample filtering, allowing the model to repeatedly focus on harder samples and gradually improve its understanding.
\begin{itemize}[nosep,wide,labelindent=0pt,labelwidth=*,align=left]
    \item \textbf{Dynamic hard sampling.}
    We adopt a multi-epoch training strategy coupled with per-epoch sample filtering to enhance learning from difficult examples.
    The model is trained over multiple epochs, and after each epoch, we exclude easy samples with an IoU greater than 0.7 that have become easy. This dynamic curriculum discourages overfitting on easy instances while ensuring consistent exposure to harder samples, ultimately promoting stronger generalization.

    \item \textbf{Cold start fine-tuning with few CoT data.} 
    For smaller models (e.g., 3B parameters), we observe that directly training with RL to generate CoT responses often results in reasoning steps that are either unintelligible or hallucinated, which impairs answer quality. 
    Additionally, the length of generated reasoning during early training stages is difficult to control, leading to an unstable training process. 
    To address these issues, we introduce a cold-start fine-tuning strategy using a small set of CoT-formatted examples that encourage grounded reasoning aligned with video content. Specifically, we guide the model to produce structured sequential captions with associated timestamps with the template as:
    \begin{equation}
\textless\mathrm{think}\textgreater \textless t_{s_1}\ \mathrm{to} \ t_{e_1}: C_1;\ t_{s_2}\ \mathrm{to} \ t_{e_2}: C_2
\textgreater\textless\mathrm{/think}\textgreater \textless\mathrm{answer}\textgreater t_s\ \mathrm{to} \ t_e\textless\mathrm{/answer}\textgreater
    \end{equation}
    where $C_i$ represent captions corresponding to video segments $[t_{s_i},t_{e_i}]$, respectively.

\end{itemize}

\begin{figure*}[t]
  \centering
  \includegraphics[width=\linewidth]{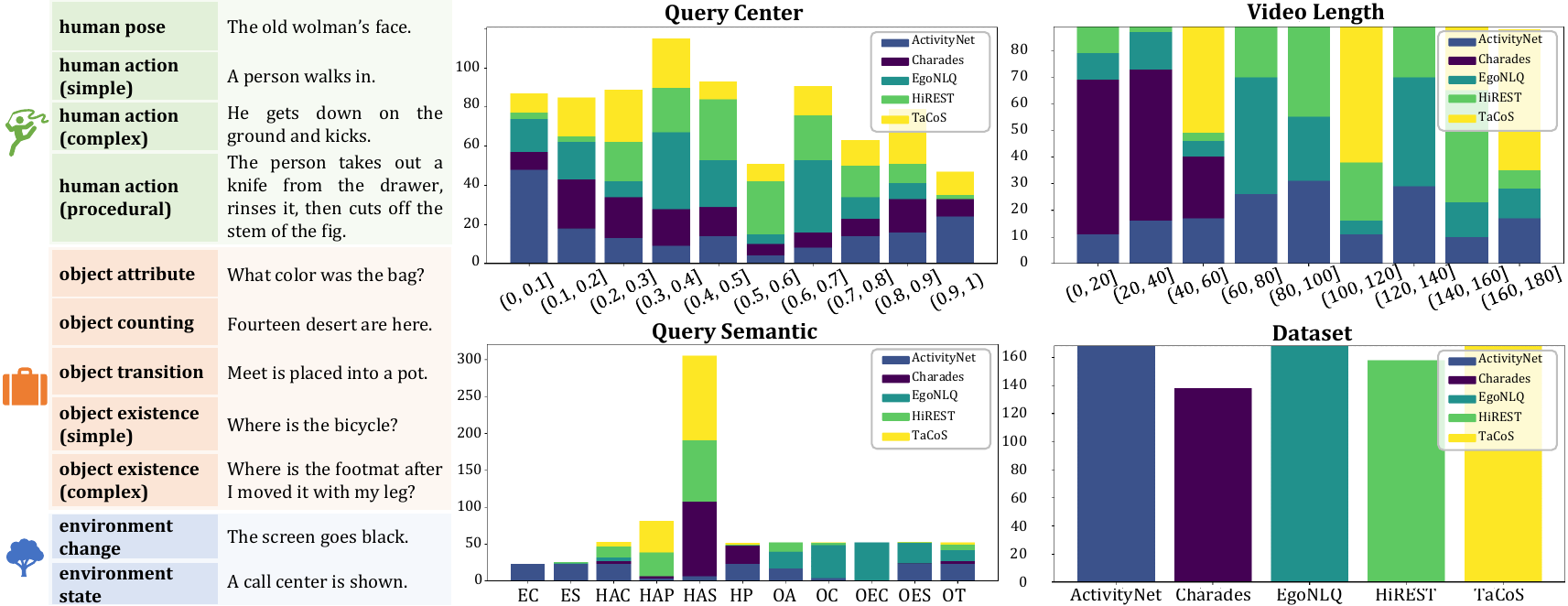}
  \caption{
  Statistics of TVGBench.
  TVGBench comprises 11 types of queries covering aspects related to humans, objects, and environments. 
  As illustrated in the figure on the right, the distributions of query center, video length, and dataset source are designed to be as balanced as possible. This balanced construction allows for a comprehensive evaluation of model performance across different dimensions, enabling fine-grained analysis along each axis during benchmarking.
  }
  \label{fig:tvgbench}
\end{figure*}

\subsection{TVGBench: Evaluation Benchmark for LVLM on Temporal Video Grounding}
\label{subsec: tvgbench}

Existing benchmarks for temporal video grounding either focus on large-scale datasets tailored for smaller models within specific domains (e.g., human activities in ActivityNet) or consist of small, limited test sets (e.g., the 2K home activity samples in Charades) typically used for LVLM evaluation due to their slower inference speed. However, these benchmarks fall short in capturing the evaluation needs of LVLMs, which, despite slower inference, exhibit strong generalization capabilities.
To bridge this gap, we introduce TVGBench, a lightweight yet comprehensive evaluation benchmark specifically designed for assessing the performance of LVLMs on temporal video grounding tasks.

\noindent\textbf{Data sources.}
To ensure a comprehensive evaluation, we construct our TVGBench by curating samples from five public benchmarks with a balanced distribution of data sources: Charades-STA~\cite{sigurdsson2016Charades}, ActivityNet-Captions~\cite{caba2015activitynet}, HiREST~\cite{zala2023hirest}, EgoNLQ~\cite{grauman2022ego4d}, and TaCoS~\cite{regneri2013tacos}.

\noindent\textbf{Benchmark statistics.}
We carefully balance the video duration, video domain, and query center, and construct query semantics in TVGBench to construct 800 instances, as shown in~\Cref{fig:tvgbench}. 

\begin{itemize}[nosep,wide,labelindent=0pt,labelwidth=*,align=left]
    \item \noindent\textbf{Distribution statistics.}
Video durations in the dataset have a balanced range from short clips up to 3 minutes long. 
To ensure temporal diversity, the center timestamps of queries are designed to be approximately uniformly distributed across the entire span of each video.

    \item \noindent\textbf{Query semantics.}
    Since the original datasets do not provide semantic labels for queries, we manually define 11 semantic categories grouped under three major types: human, object, and environment. We leverage DeepSeek-V3~\cite{deepseekv3} to annotate the semantic type of each query and ensure approximate balance across these categories.
    While most categories are evenly represented, the Human Action Simple (HAS) category is slightly overrepresented due to inherent dataset bias (simple indoor actions are more common). In such cases, we prioritize achieving a balance across datasets from different domains while maintaining semantic diversity, accepting a skew in HAS.
\end{itemize}

\section{Experiments}
\label{sec:exp}
We first present our experimental setup in~\Cref{subsec: exp_setups}.
Then, we evaluate our model from three key perspectives:
(1) Comparison with state-of-the-art methods in~\Cref{subsec: comp_with_sota}: We evaluate our model across multiple TVG benchmarks to assess its performance against existing approaches;
(2) Ablation studies and analyses in~\Cref{subsec: exp_ablation}: We examine the individual contributions of each component in our framework to better understand their roles in overall performance. We also compare RL and SFT strategies across TVG, short video QA, and long video QA tasks.

\subsection{Experimental Setup}

\label{subsec: exp_setups}
\noindent\textbf{Benchmarks.}
We evaluate our model on a wide range of benchmarks covering both temporal video grounding and general video understanding tasks, including:
(1) \textit{Charades-STA}~\cite{sigurdsson2016Charades} contains 6,672 long videos capturing indoor human activities. The official split for the TVG task includes 12,408 clip-query pairs for training and 3,720 for testing.
(2) \textit{ActivityNet}~\cite{caba2015activitynet} comprises 20K long videos with an average of 3.65 clip-query pairs per video. 
Following previous work in fine-tuning setting~\cite{zhang2021ms2dtan,jang2023knowing} for the TVG task, we use the standard dataset splits with 37,421 training, 17,505 validation, and 17,031 test samples.
(3) \textit{MVBench}~\cite{li2024mvbench} is a short video QA benchmark focused on temporal reasoning. It includes 4K QA pairs for 20 types of tasks.
(4) \textit{TempCompass}~\cite{liu2024tempcompass} assesses fine-grained temporal understanding with 410 short videos. We use all multi-choice QA tasks except for the video captioning task.  
(5) \textit{EgoSchema}~\cite{mangalam2023egoschema} features 5K egocentric video clips, each approximately 3 minutes long, with temporally demanding QA pairs.
(6) \textit{VideoMME}~\cite{fu2024videomme} is a general video QA benchmark covering diverse domains. It contains 2.7K QA samples over videos of varied lengths, ranging from 11 seconds to 1 hour. We use the long video split for evaluation.

\noindent\textbf{Implementation details.} 
Unless otherwise specified, we use Qwen2.5-VL-7B~\cite{Qwen2.5-VL} as the base model.
To strike a balance between training efficiency and memory consumption, we sample video frames at 2 FPS and adaptively resize each video input to contain approximately 2.8 million pixels.
For instance, a 50-second video yields 100 frames, each with a resolution of roughly 96$\times$96$\times$3.
During the reinforcement fine-tuning phase, we train for 5 epochs using a batch size of 8 and select the final checkpoint for evaluation.
For fine-tuning on downstream benchmarks, we train for 2 epochs.
All experiments are conducted on a cluster with 8$\times$NVIDIA A100 GPUs.

\noindent\textbf{Evaluation metrics.}
For TVG, following~\cite{ren2024timechat,zeng2025timesuite}, we adopt the ``R1@m'' evaluation protocol to compare with state-of-the-art models, which computes the percentage of samples where the top-1 predicted segment has an IoU greater than a threshold $m$, with $m\in\{0.3, 0.5, 0.7\}$.
For brevity, we also adopt mIoU, which calculates the average IoU on all testing data as an alternative metric.
For video QA, we report accuracy as the evaluation metric.

\begin{table*}[t]
\centering
\caption{
Performance of temporal video grounding on Charades-STA, ActivityNet, and TVGBench.  
The methods marked in \textcolor{gray}{gray$^*$} represent fine-tuning on corresponding benchmarks, while those in black indicate zero-shot settings.
We compare our Time-R1 against existing 7B open-source LVLMs, as well as state-of-the-art VLP models.
}
\setlength{\belowcaptionskip}{3pt}%
\scalebox{0.75}{
\begin{tabular}{cl|cccc@{}|cccc@{}|ccc}
\toprule
\multirow{2}{*}{Type} & \multirow{2}{*}{Method} & \multicolumn{3}{c}{Charades-STA} && \multicolumn{3}{c}{ActivityNet} && \multicolumn{3}{c}{TVGBench}\\
 & & {\fontsize{8.4}{10}\selectfont R1@0.3} & {\fontsize{8.4}{10}\selectfont R1@0.5} & {\fontsize{8.4}{10}\selectfont R1@0.7} && {\fontsize{8.4}{10}\selectfont R1@0.3} & {\fontsize{8.4}{10}\selectfont R1@0.5} & {\fontsize{8.4}{10}\selectfont R1@0.7} && {\fontsize{8.4}{10}\selectfont R1@0.3} & {\fontsize{8.4}{10}\selectfont R1@0.5} & {\fontsize{8.4}{10}\selectfont R1@0.7}\\
\midrule
\multirow{4}{*}{\textcolor{gray}{VLP}} 
& \textcolor{gray}{2D-TAN$^*$~\cite{zhang20192DTAN}} & 
\textcolor{gray}{57.3} & \textcolor{gray}{45.8} & \textcolor{gray}{27.9} && \textcolor{gray}{60.4} & \textcolor{gray}{43.4} & \textcolor{gray}{25.0} && - & - & -  \\ 
& \textcolor{gray}{UniVTG$^*$~\cite{lin2023univtg}} & 
\textcolor{gray}{72.6} & \textcolor{gray}{60.2} & \textcolor{gray}{38.6} && \textcolor{gray}{56.1} & \textcolor{gray}{43.4} & \textcolor{gray}{24.3} && - & - & -  \\ 
 & \textcolor{gray}{SSRN$^*$~\cite{ssrn}} & 
- & \textcolor{gray}{65.5} & \textcolor{gray}{42.6} && - & \textcolor{gray}{54.5} & \textcolor{gray}{33.2} && - & - & -  \\ 
 & \textcolor{gray}{SnAG$^*$~\cite{CVPR2024SnAG}} & 
- & \textcolor{gray}{64.6} & \textcolor{gray}{46.2} && - & \textcolor{gray}{48.6} & \textcolor{gray}{30.6} && - & - & - \\ 
 & \textcolor{gray}{EaTR$^*$~\cite{jang2023knowing}} & 
- & \textcolor{gray}{68.4} & \textcolor{gray}{44.9}  && - & \textcolor{gray}{58.2} & \textcolor{gray}{37.6} && - & - & - \\ 
\midrule
 & Gemini-2.5-Pro~\cite{Gemini2.5} & - & - & - && - & - & - && 39.1 & 24.4 & 12.8  \\
\midrule
\multirow{9}{*}{SFT} & ChatVTG~\cite{qu2024chatvtg} & 
52.7 & 33.0 & 15.9 && 40.7 & 22.5 & 9.4 && - & - & - \\  
 & TimeChat~\cite{ren2024timechat} & 
- & 32.2 & 13.4 && 36.2 & 20.2 & 9.5 && 22.4 & 11.9 & 5.3 \\  
 & HawkEye~\cite{wang2024hawkeye} & 
50.6 & 31.4 & 14.5  && 49.1 & 29.3 & 10.7 && - & - & - \\  
 & VTimeLLM~\cite{huang2024vtimellm} & 
51.0 & 27.5 & 11.4 && 44.0 & 27.8 & 14.3 && - & - & - \\
 & TimeSuite~\cite{zeng2025timesuite} & 
69.9 & 48.7 & 24.0 && - & - & - && 31.1 & 18.0 & 8.9 \\ 
 & VideoChat-Flash~\cite{li2024videochatflash} & 
74.5 & 53.1 & 27.6 && - & - & - && 32.8 & 19.8 & 10.4 \\  
 & TRACE~\cite{guo2024trace} & 
- & 40.3 & 19.4  && - & \textcolor{gray}{37.7} & \textcolor{gray}{24.0} && 37.0 & 25.5 & 14.6 \\
 & \textcolor{gray}{HawkEye$^*$~\cite{wang2024hawkeye}} & 
\textcolor{gray}{72.5} & \textcolor{gray}{58.3} & \textcolor{gray}{28.8}  && \textcolor{gray}{55.9} & \textcolor{gray}{34.7} & \textcolor{gray}{17.9} && - & - & - \\  
& \textcolor{gray}{TimeSuite$^*$~\cite{zeng2025timesuite}} & 
\textcolor{gray}{79.4} & \textcolor{gray}{67.1} & \textcolor{gray}{43.0} && - & - & - && - & - & - \\ 

\midrule
\multirow{2}{*}{RL} & \cellcolor{mygray} Time-R1 (ours) & 
\cellcolor{mygray}\textbf{78.1} & \cellcolor{mygray}\textbf{60.8} & \cellcolor{mygray}\textbf{35.3}&\cellcolor{mygray}& \cellcolor{mygray}\textbf{58.6} & \cellcolor{mygray}\textbf{39.0} & \cellcolor{mygray}\textbf{21.4}&\cellcolor{mygray}& \cellcolor{mygray}\textbf{41.8} & \cellcolor{mygray} \textbf{29.4}
 & \cellcolor{mygray}\textbf{16.4} \\  
& \cellcolor{mygray} \textcolor{gray}{Time-R1 (ours)$^*$}& 
\cellcolor{mygray}\textcolor{gray}{82.8} & \cellcolor{mygray}\textcolor{gray}{72.2} & \cellcolor{mygray}\textcolor{gray}{50.1} &\cellcolor{mygray} & \cellcolor{mygray}\textcolor{gray}{73.3} & \cellcolor{mygray}\textcolor{gray}{55.6} & \cellcolor{mygray}\textcolor{gray}{34.0} &\cellcolor{mygray}& \cellcolor{mygray} \textcolor{gray}{-} & \cellcolor{mygray} \textcolor{gray}{-} & \cellcolor{mygray} \textcolor{gray}{-} \\  
\bottomrule
\end{tabular}}
\label{tab:comp_TVG}
\end{table*}

\begin{figure*}[t]
  \centering
  \includegraphics[width=\linewidth]{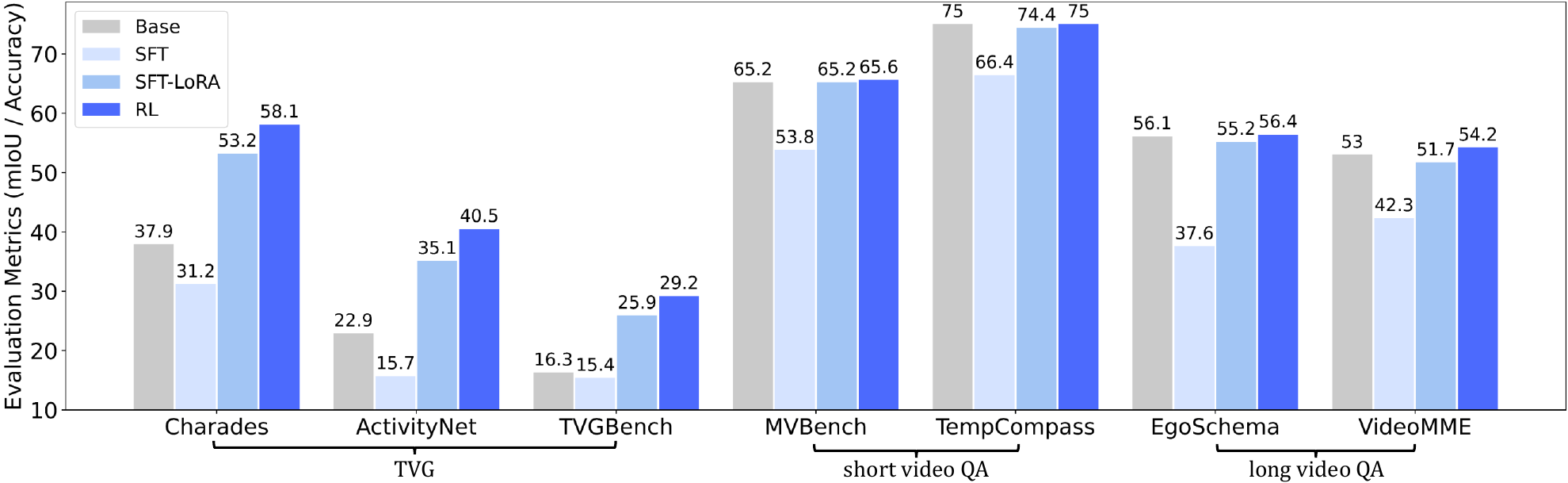}
  \caption{Comparison between post-training paradigms across various tasks, including temporal video grounding, short video QA, and long video QA. Both ``SFT'' and ``RL'' full-finetune the LLM, while ``SFT-LoRA'' denotes finetuning the LLM with LoRA~\cite{hu2022lora}. The ``Base'' is Qwen2.5-VL-7B.
  }
  \label{fig:rl_vs_sft}
\end{figure*}

\subsection{Comparison with State-of-the-Art}
\label{subsec: comp_with_sota}

We compare Time-R1 with state-of-the-art TVG methods, including both traditional video-language pre-training models (VLP)  and recent large video-language models fine-tuned via SFT. 

\noindent\textbf{Time-R1 surpasses SFT-based models in zero-shot settings.}
As shown in~\Cref{tab:comp_TVG}, in the zero-shot setting, Time-R1 demonstrates strong performance, outperforming SFT-based models that rely on large-scale supervision. 
Despite using only 2.5K samples for RL, Time-R1 achieves leading results across multiple benchmarks.
For example, on Charades-STA, Time-R1 attains an R1@0.7 score of 35.3, outperforming VideoChat-Flash (27.6) and TimeSuite (24.0).
On ActivityNet, it achieves R1@0.7 score of 21.4, surpassing VTimeLLM (14.3) and TimeChat (10.7).
On TVGBench, it scores 16.4, outperforming TRACE (14.6) and Gemini-2.5-Pro (12.8).

\begin{wraptable}{r}{0.5\textwidth} 
\centering
\caption{Ablation of Time-R1-7B trainning.
GF, ME, SF refers to Gaussian Filtering, Multi-Epoch, and Sample Filtering per epoch, respectively.
}
\vspace{-0.1cm}
\setlength{\belowcaptionskip}{3pt}%
\scalebox{0.75}{
\begin{tabular}{c|cc|cc|ccc}
\toprule
& \multirow{2}{*}{tIoU} & \multirow{2}{*}{GF} & \multirow{2}{*}{ME} & \multirow{2}{*}{SF} & \multicolumn{3}{c}{TVGBench}\\
& &&&& {\fontsize{8.4}{10}\selectfont R1@0.3} & {\fontsize{8.4}{10}\selectfont R1@0.5} & {\fontsize{8.4}{10}\selectfont R1@0.7}\\
\midrule
1 & \ding{55} & \ding{55} & \ding{55} & \ding{55} & 38.0 & 24.8 & 13.2 \\
2 & \ding{51} & \ding{55} & \ding{55} & \ding{55} & 36.0 & 23.6 & 12.9 \\  
3 & \ding{55} & \ding{51} & \ding{55} & \ding{55} & 37.2 & 25.0 & 13.4 \\   
4 & \ding{55} & \ding{55} & \ding{51} & \ding{55} & 39.9 & 26.0 & 14.2 \\   
5 & \ding{51} & \ding{51} & \ding{55} & \ding{55} & 38.4 & 25.6 & 14.1 \\   
6 & \ding{51} & \ding{55} & \ding{51} & \ding{55} & 39.4 & 26.5 & 16.4 \\   
7 & \ding{51} & \ding{51} & \ding{51} & \ding{55} & 41.6 & 28.5 & 15.6 \\   
8 & \ding{51} & \ding{51} & \ding{51} & \ding{51} & 41.8 & 29.4 & 16.4 \\   
\bottomrule
\end{tabular}
}
\vspace{-0.1cm}
\label{tab:ablation}
\end{wraptable}

\noindent\textbf{Time-R1$^*$ outperforms all SFT-based LVLMs and many traditional VLP-based models.}
Time-R1$^*$ consistently outperforms both traditional VLP-based and SFT-based models on the TVG task. 
On Charades-STA, it exceeds EaTR  
and fine-tuned TimeSuite    
 by 3.4-7.1 percentage points across R1@0.3 to R1@0.7. 
Notably, Time-R1$^*$ surpasses TimeSuite despite using far fewer RL samples compared to TimeSuite’s 349K SFT examples.
On the more challenging ActivityNet dataset, Time-R1$^*$ also outperforms SSRN~\cite{ssrn} and TRACE~\cite{guo2024trace}, achieving significant gains across key metrics.

\subsection{Ablation Study}
\label{subsec: exp_ablation}
We conduct a detailed ablation on the Time-R1-7B model to investigate the contribution of various training strategies. 

\noindent\textbf{Utility of TimeRFT and Time-R1 components.}
As shown in Table~\ref{tab:ablation},  both Gaussian Filtering (GF) and Multi-Epoch training (ME) individually improve performance, with ME yielding a more substantial gain, improving from R1@0.7 of 13.2 in row 1 to 14.2 in row 4.
Notably, the combination of tIoU supervision and ME (Row 6) leads to a significant boost across all metrics. 
As more components are added, GF and ME (Row 7), followed by Sample Filtering (SF) in Row 8, the performance continues to improve, ultimately reaching R1@0.5 of 29.4 and R1@0.7 of 16.4.

\begin{wrapfigure}{r}{0.4\textwidth} 
  \centering
    \includegraphics[width=\linewidth]{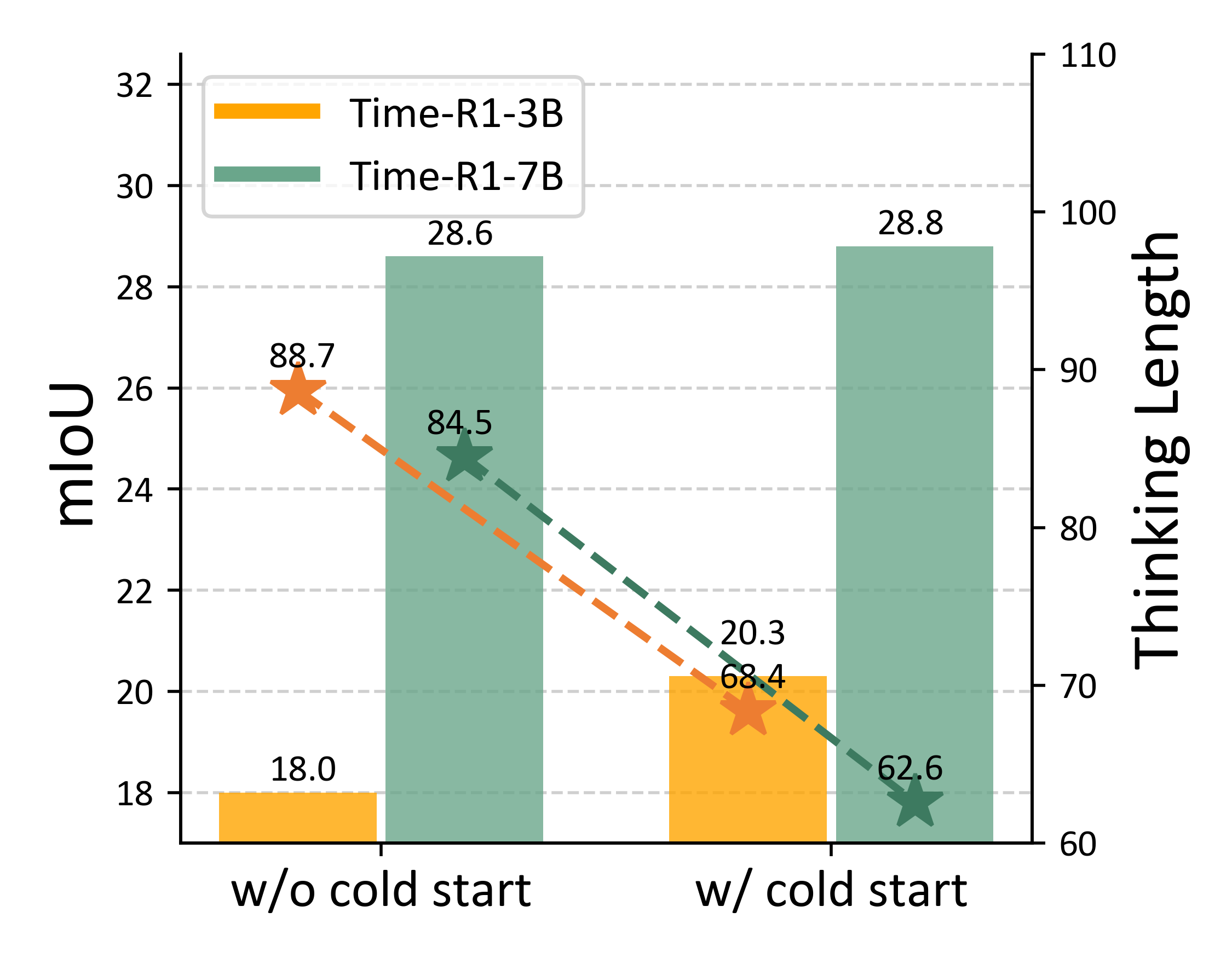}
  \caption{Impact of SFT-based cold start on IoU performance and thinking token count, with token counts marked by \(\star\) on dashed lines.}
  \label{fig:motivation}
\end{wrapfigure}

\noindent\textbf{Generalization of RL vs. SFT.}
As shown in~\Cref{fig:rl_vs_sft}, when both RL and SFT perform full fine-tuning of the LLM using a small amount of data, SFT significantly degrades the model's generalization ability across all tasks, whereas RL consistently preserves generalization. 
While SFT can alleviate this issue by adopting LoRA-based fine-tuning (SFT-LoRA), RL still demonstrates stronger generalization and additionally improves performance on video QA tasks, whereas SFT-LoRA leads to a decline. 
For example, on ActivityNet, RL improves the mIoU from 16.3 to 29.2, while SFT degrades it to 15.4, and SFT-LoRA improves it to 25.9. 
On the VideoMMe video QA benchmark, RL increases the performance from 53.0 to 54.2, whereas SFT-LoRA decreases it to 51.7.

\noindent\textbf{Impact of cold start.} 
As shown in~\Cref{fig:motivation}, cold start boosts the performance of the 3B model and reduces the number of thinking tokens for both models. 
We attribute this to the cold start's function to suppress hallucinations, which tend to be more prevalent in weaker models.

\section{Conclusion}
In this work, we present Time-R1, a reinforcement learning based post-training framework that significantly improves the generalization of Large Vision-Language Models for Temporal Video Grounding. Unlike prior methods that rely on large-scale supervised fine-tuning, Time-R1 leverages a verifiable reward signal to unlock strong temporal reasoning from pretrained models using limited data.
Our contributions include:
(1) Time-R1, a reasoning-guided post-training framework that enhances TVG via RL;
(2) TimeRFT, a curated dataset and training strategy that fosters temporal grounding;
(3) TVGBench, a small yet comprehensive benchmark for evaluating LVLMs on TVG.
Extensive experiments show that Time-R1 achieves SoTA performance across TVG benchmarks in both zero-shot and fine-tuned settings, surpassing prior LVLMs and traditional VLP-based models, while also improving general video understanding. 
We hope this work inspires future directions in data-efficient and generalizable video-language understanding via reinforcement learning.

\clearpage

{
    \small
    \bibliographystyle{plain}
    \bibliography{main}

\begin{thebibliography}{10}

\bibitem{deepseekr1}
Deepseek-r1: Incentivizing reasoning capability in llms via reinforcement learning.
\newblock {\em arXiv preprint arXiv:2501.12948}, 2025.

\bibitem{afouras2023htstep}
Triantafyllos Afouras, Effrosyni Mavroudi, Tushar Nagarajan, Huiyu Wang, and Lorenzo Torresani.
\newblock Ht-step: Aligning instructional articles with how-to videos.
\newblock {\em Advances in Neural Information Processing Systems}, 36:50310--50326, 2023.

\bibitem{anne2017didemo}
Lisa Anne~Hendricks, Oliver Wang, Eli Shechtman, Josef Sivic, Trevor Darrell, and Bryan Russell.
\newblock Localizing moments in video with natural language.
\newblock In {\em Proceedings of the IEEE international conference on computer vision}, pages 5803--5812, 2017.

\bibitem{Qwen2.5-VL}
Shuai Bai, Keqin Chen, Xuejing Liu, Jialin Wang, Wenbin Ge, Sibo Song, Kai Dang, Peng Wang, Shijie Wang, Jun Tang, Humen Zhong, Yuanzhi Zhu, Mingkun Yang, Zhaohai Li, Jianqiang Wan, Pengfei Wang, Wei Ding, Zheren Fu, Yiheng Xu, Jiabo Ye, Xi~Zhang, Tianbao Xie, Zesen Cheng, Hang Zhang, Zhibo Yang, Haiyang Xu, and Junyang Lin.
\newblock Qwen2.5-vl technical report.
\newblock {\em arXiv preprint arXiv:2502.13923}, 2025.

\bibitem{caba2015activitynet}
Fabian Caba~Heilbron, Victor Escorcia, Bernard Ghanem, and Juan Carlos~Niebles.
\newblock Activitynet: A large-scale video benchmark for human activity understanding.
\newblock In {\em Proceedings of the ieee conference on computer vision and pattern recognition}, pages 961--970, 2015.

\bibitem{Carreira_2017_CVPR_I3D}
Joao Carreira and Andrew Zisserman.
\newblock Quo vadis, action recognition? a new model and the kinetics dataset.
\newblock In {\em proceedings of the IEEE Conference on Computer Vision and Pattern Recognition}, pages 6299--6308, 2017.

\bibitem{chen2025r1v}
Liang Chen, Lei Li, Haozhe Zhao, Yifan Song, and Vinci.
\newblock R1-v: Reinforcing super generalization ability in vision-language models with less than \$3.
\newblock \url{https://github.com/Deep-Agent/R1-V}, 2025.

\bibitem{instructblip}
Wenliang Dai, Junnan Li, Dongxu Li, Anthony Meng~Huat Tiong, Junqi Zhao, Weisheng Wang, Boyang Li, Pascale Fung, and Steven Hoi.
\newblock Instructblip: Towards general-purpose vision-language models with instruction tuning, 2023.

\bibitem{darrell1993space}
Trevor Darrell and Alex Pentland.
\newblock Space-time gestures.
\newblock In {\em Proceedings of IEEE Conference on Computer Vision and Pattern Recognition}, pages 335--340. IEEE, 1993.

\bibitem{Gemini2.5}
Google DeepMind.
\newblock Gemini 2.5: Our most intelligent ai model.
\newblock {\em Google DeepMind}, 2025.
\newblock Model ID: gemini-2.5-pro-preview-03-25.

\bibitem{deepseekv3}
DeepSeek-AI.
\newblock Deepseek llm: Scaling open-source language models with longtermism.
\newblock {\em arXiv preprint arXiv:2401.02954}, 2024.

\bibitem{devlin2019bert}
Jacob Devlin, Ming{-}Wei Chang, Kenton Lee, and Kristina Toutanova.
\newblock {BERT:} pre-training of deep bidirectional transformers for language understanding.
\newblock In {\em Proceedings of the 2019 Conference of the North American Chapter of the Association for Computational Linguistics}, pages 4171--4186, 2019.

\bibitem{fu2024videomme}
Chaoyou Fu, Yuhan Dai, Yongdong Luo, Lei Li, Shuhuai Ren, Renrui Zhang, Zihan Wang, Chenyu Zhou, Yunhang Shen, Mengdan Zhang, et~al.
\newblock Video-mme: The first-ever comprehensive evaluation benchmark of multi-modal llms in video analysis.
\newblock In {\em Proceedings of the IEEE/CVF Conference on Computer Vision and Pattern Recognition}, 2025.

\bibitem{gaidon2013temporal}
Adrien Gaidon, Zaid Harchaoui, and Cordelia Schmid.
\newblock Temporal localization of actions with actoms.
\newblock {\em IEEE transactions on pattern analysis and machine intelligence}, 35(11):2782--2795, 2013.

\bibitem{gao2017tall}
Jiyang Gao, Chen Sun, Zhenheng Yang, and Ram Nevatia.
\newblock Tall: Temporal activity localization via language query.
\newblock In {\em Proceedings of the IEEE international conference on computer vision}, pages 5267--5275, 2017.

\bibitem{grauman2022ego4d}
Kristen Grauman, Andrew Westbury, Eugene Byrne, Zachary Chavis, Antonino Furnari, Rohit Girdhar, Jackson Hamburger, Hao Jiang, Miao Liu, Xingyu Liu, et~al.
\newblock Ego4d: Around the world in 3,000 hours of egocentric video.
\newblock In {\em Proceedings of the IEEE/CVF Conference on Computer Vision and Pattern Recognition}, pages 18995--19012, 2022.

\bibitem{guo2025vtgllm}
Yongxin Guo, Jingyu Liu, Mingda Li, Dingxin Cheng, Xiaoying Tang, Dianbo Sui, Qingbin Liu, Xi~Chen, and Kevin Zhao.
\newblock Vtg-llm: Integrating timestamp knowledge into video llms for enhanced video temporal grounding.
\newblock In {\em Proceedings of the AAAI Conference on Artificial Intelligence}, volume~39, pages 3302--3310, 2025.

\bibitem{guo2024trace}
Yongxin Guo, Jingyu Liu, Mingda Li, Qingbin Liu, Xi~Chen, and Xiaoying Tang.
\newblock Trace: Temporal grounding video llm via causal event modeling.
\newblock {\em arXiv preprint arXiv:2410.05643}, 2024.

\bibitem{hannan2024revisionllm}
Tanveer Hannan, Md~Mohaiminul Islam, Jindong Gu, Thomas Seidl, and Gedas Bertasius.
\newblock Revisionllm: Recursive vision-language model for temporal grounding in hour-long videos.
\newblock In {\em Proceedings of the IEEE/CVF Conference on Computer Vision and Pattern Recognition}, 2025.

\bibitem{hu2022lora}
Edward~J Hu, Yelong Shen, Phillip Wallis, Zeyuan Allen-Zhu, Yuanzhi Li, Shean Wang, Lu~Wang, Weizhu Chen, et~al.
\newblock Lora: Low-rank adaptation of large language models.
\newblock {\em ICLR}, 1(2):3, 2022.

\bibitem{huang2024vtimellm}
Bin Huang, Xin Wang, Hong Chen, Zihan Song, and Wenwu Zhu.
\newblock Vtimellm: Empower llm to grasp video moments.
\newblock In {\em Proceedings of the IEEE/CVF Conference on Computer Vision and Pattern Recognition}, pages 14271--14280, 2024.

\bibitem{jang2023knowing}
Jinhyun Jang, Jungin Park, Jin Kim, Hyeongjun Kwon, and Kwanghoon Sohn.
\newblock Knowing where to focus: Event-aware transformer for video grounding.
\newblock In {\em Proceedings of the IEEE/CVF International Conference on Computer Vision}, pages 13846--13856, 2023.

\bibitem{kwon2023vllm}
Woosuk Kwon, Zhuohan Li, Siyuan Zhuang, Ying Sheng, Lianmin Zheng, Cody~Hao Yu, Joseph~E. Gonzalez, Hao Zhang, and Ion Stoica.
\newblock Efficient memory management for large language model serving with pagedattention.
\newblock In {\em Proceedings of the ACM SIGOPS 29th Symposium on Operating Systems Principles}, 2023.

\bibitem{laptev2007retrieving}
Ivan Laptev and Patrick P{\'e}rez.
\newblock Retrieving actions in movies.
\newblock In {\em 2007 IEEE 11th International Conference on Computer Vision}, pages 1--8. IEEE, 2007.

\bibitem{li2025imove}
Jiaze Li, Yaya Shi, Zongyang Ma, Haoran Xu, Feng Cheng, Huihui Xiao, Ruiwen Kang, Fan Yang, Tingting Gao, and Di~Zhang.
\newblock imove: Instance-motion-aware video understanding.
\newblock {\em arXiv preprint arXiv:2502.11594}, 2025.

\bibitem{li2024mvbench}
Kunchang Li, Yali Wang, Yinan He, Yizhuo Li, Yi~Wang, Yi~Liu, Zun Wang, Jilan Xu, Guo Chen, Ping Luo, et~al.
\newblock Mvbench: A comprehensive multi-modal video understanding benchmark.
\newblock In {\em Proceedings of the IEEE/CVF Conference on Computer Vision and Pattern Recognition}, pages 22195--22206, 2024.

\bibitem{li2024videochatflash}
Xinhao Li, Yi~Wang, Jiashuo Yu, Xiangyu Zeng, Yuhan Zhu, Haian Huang, Jianfei Gao, Kunchang Li, Yinan He, Chenting Wang, Yu~Qiao, Yali Wang, and Limin Wang.
\newblock Videochat-flash: Hierarchical compression for long-context video modeling.
\newblock {\em arXiv preprint arXiv:2501.00574}, 2024.

\bibitem{liao2025improved}
Zhenyi Liao, Qingsong Xie, Yanhao Zhang, Zijian Kong, Haonan Lu, Zhenyu Yang, and Zhijie Deng.
\newblock Improved visual-spatial reasoning via r1-zero-like training.
\newblock {\em arXiv preprint arXiv:2504.00883}, 2025.

\bibitem{lin2022egovlp}
Kevin~Qinghong Lin, Jinpeng Wang, Mattia Soldan, Michael Wray, Rui Yan, Eric~Z Xu, Difei Gao, Rong-Cheng Tu, Wenzhe Zhao, Weijie Kong, et~al.
\newblock Egocentric video-language pretraining.
\newblock {\em Advances in Neural Information Processing Systems}, 35:7575--7586, 2022.

\bibitem{lin2023univtg}
Kevin~Qinghong Lin, Pengchuan Zhang, Joya Chen, Shraman Pramanick, Difei Gao, Alex~Jinpeng Wang, Rui Yan, and Mike~Zheng Shou.
\newblock Univtg: Towards unified video-language temporal grounding.
\newblock In {\em Proceedings of the IEEE/CVF International Conference on Computer Vision}, pages 2794--2804, 2023.

\bibitem{liu2024tempcompass}
Yuanxin Liu, Shicheng Li, Yi~Liu, Yuxiang Wang, Shuhuai Ren, Lei Li, Sishuo Chen, Xu~Sun, and Lu~Hou.
\newblock Tempcompass: Do video llms really understand videos?
\newblock {\em arXiv preprint arXiv:2403.00476}, 2024.

\bibitem{liu2025visualrft}
Ziyu Liu, Zeyi Sun, Yuhang Zang, Xiaoyi Dong, Yuhang Cao, Haodong Duan, Dahua Lin, and Jiaqi Wang.
\newblock Visual-rft: Visual reinforcement fine-tuning.
\newblock {\em arXiv preprint arXiv:2503.01785}, 2025.

\bibitem{mangalam2023egoschema}
Karttikeya Mangalam, Raiymbek Akshulakov, and Jitendra Malik.
\newblock Egoschema: A diagnostic benchmark for very long-form video language understanding.
\newblock {\em Advances in Neural Information Processing Systems}, 36:46212--46244, 2023.

\bibitem{margolis2023walk}
Gabriel~B Margolis and Pulkit Agrawal.
\newblock Walk these ways: Tuning robot control for generalization with multiplicity of behavior.
\newblock In {\em Conference on Robot Learning}, pages 22--31. PMLR, 2023.

\bibitem{meng2025mmeureka}
Fanqing Meng, Lingxiao Du, Zongkai Liu, Zhixiang Zhou, Quanfeng Lu, Daocheng Fu, Botian Shi, Wenhai Wang, Junjun He, Kaipeng Zhang, et~al.
\newblock Mm-eureka: Exploring visual aha moment with rule-based large-scale reinforcement learning.
\newblock {\em arXiv preprint arXiv:2503.07365}, 2025.

\bibitem{miech2019howto100m}
Antoine Miech, Dimitri Zhukov, Jean-Baptiste Alayrac, Makarand Tapaswi, Ivan Laptev, and Josef Sivic.
\newblock Howto100m: Learning a text-video embedding by watching hundred million narrated video clips.
\newblock In {\em Proceedings of the IEEE/CVF international conference on computer vision}, pages 2630--2640, 2019.

\bibitem{CVPR2024SnAG}
Fangzhou Mu, Sicheng Mo, and Yin Li.
\newblock Snag: Scalable and accurate video grounding.
\newblock {\em 2024 IEEE/CVF Conference on Computer Vision and Pattern Recognition (CVPR)}, pages 18930--18940, 2024.

\bibitem{oncescu2021queryd}
Andreea-Maria Oncescu, Joao~F Henriques, Yang Liu, Andrew Zisserman, and Samuel Albanie.
\newblock Queryd: A video dataset with high-quality text and audio narrations.
\newblock In {\em ICASSP 2021-2021 IEEE International Conference on Acoustics, Speech and Signal Processing (ICASSP)}, pages 2265--2269. IEEE, 2021.

\bibitem{openaio1}
OpenAI.
\newblock Openai o1, 2024.

\bibitem{ouyang2022rlhf}
Long Ouyang, Jeffrey Wu, Xu~Jiang, Diogo Almeida, Carroll Wainwright, Pamela Mishkin, Chong Zhang, Sandhini Agarwal, Katarina Slama, Alex Ray, et~al.
\newblock Training language models to follow instructions with human feedback.
\newblock {\em Advances in neural information processing systems}, 35:27730--27744, 2022.

\bibitem{qu2024chatvtg}
Mengxue Qu, Xiaodong Chen, Wu~Liu, Alicia Li, and Yao Zhao.
\newblock Chatvtg: Video temporal grounding via chat with video dialogue large language models.
\newblock In {\em Proceedings of the IEEE/CVF Conference on Computer Vision and Pattern Recognition}, pages 1847--1856, 2024.

\bibitem{radford2021clip}
Alec Radford, Jong~Wook Kim, Chris Hallacy, Aditya Ramesh, Gabriel Goh, Sandhini Agarwal, Girish Sastry, Amanda Askell, Pamela Mishkin, Jack Clark, et~al.
\newblock Learning transferable visual models from natural language supervision.
\newblock In {\em International conference on machine learning}, pages 8748--8763. PmLR, 2021.

\bibitem{regneri2013tacos}
Michaela Regneri, Marcus Rohrbach, Dominikus Wetzel, Stefan Thater, Bernt Schiele, and Manfred Pinkal.
\newblock Grounding action descriptions in videos.
\newblock {\em Transactions of the Association for Computational Linguistics}, 1:25--36, 2013.

\bibitem{ren2024timechat}
Shuhuai Ren, Linli Yao, Shicheng Li, Xu~Sun, and Lu~Hou.
\newblock Timechat: A time-sensitive multimodal large language model for long video understanding.
\newblock In {\em Proceedings of the IEEE/CVF Conference on Computer Vision and Pattern Recognition}, pages 14313--14323, 2024.

\bibitem{grpo}
Zhihong Shao, Peiyi Wang, Qihao Zhu, Runxin Xu, Junxiao Song, Xiao Bi, Haowei Zhang, Mingchuan Zhang, Y.~K. Li, Y.~Wu, and Daya Guo.
\newblock Deepseekmath: Pushing the limits of mathematical reasoning in open language models.
\newblock {\em arXiv preprint arXiv:2402.03300}, 2024.

\bibitem{sigurdsson2016Charades}
Gunnar~A Sigurdsson, G{\"u}l Varol, Xiaolong Wang, Ali Farhadi, Ivan Laptev, and Abhinav Gupta.
\newblock Hollywood in homes: Crowdsourcing data collection for activity understanding.
\newblock In {\em Proceedings of the European Conference on Computer Vision (ECCV)}, 2016.

\bibitem{silver2017alphazero}
David Silver, Thomas Hubert, Julian Schrittwieser, Ioannis Antonoglou, Matthew Lai, Arthur Guez, Marc Lanctot, Laurent Sifre, Dharshan Kumaran, Thore Graepel, et~al.
\newblock Mastering chess and shogi by self-play with a general reinforcement learning algorithm.
\newblock {\em arXiv preprint arXiv:1712.01815}, 2017.

\bibitem{tan2025reasonrft}
Huajie Tan, Yuheng Ji, Xiaoshuai Hao, Minglan Lin, Pengwei Wang, Zhongyuan Wang, and Shanghang Zhang.
\newblock Reason-rft: Reinforcement fine-tuning for visual reasoning.
\newblock {\em arXiv preprint arXiv:2503.20752}, 2025.

\bibitem{wang2023internvid}
Yi~Wang, Yinan He, Yizhuo Li, Kunchang Li, Jiashuo Yu, Xin Ma, Xinhao Li, Guo Chen, Xinyuan Chen, Yaohui Wang, et~al.
\newblock Internvid: A large-scale video-text dataset for multimodal understanding and generation.
\newblock {\em arXiv preprint arXiv:2307.06942}, 2023.

\bibitem{wang2024hawkeye}
Yueqian Wang, Xiaojun Meng, Jianxin Liang, Yuxuan Wang, Qun Liu, and Dongyan Zhao.
\newblock Hawkeye: Training video-text llms for grounding text in videos, 2024.

\bibitem{transformers_lib}
Thomas Wolf, Lysandre Debut, Victor Sanh, Julien Chaumond, Clement Delangue, Anthony Moi, Pierric Cistac, Tim Rault, Rémi Louf, Morgan Funtowicz, Joe Davison, Sam Shleifer, Patrick von Platen, Clara Ma, Yacine Jernite, Julien Plu, Canwen Xu, Teven~Le Scao, Sylvain Gugger, Mariama Drame, Quentin Lhoest, and Alexander~M. Rush.
\newblock Transformers: State-of-the-art natural language processing.
\newblock In {\em Proceedings of the 2020 Conference on Empirical Methods in Natural Language Processing: System Demonstrations}, pages 38--45, Online, October 2020. Association for Computational Linguistics.

\bibitem{wu2024numberit}
Yongliang Wu, Xinting Hu, Yuyang Sun, Yizhou Zhou, Wenbo Zhu, Fengyun Rao, Bernt Schiele, and Xu~Yang.
\newblock Number it: Temporal grounding videos like flipping manga.
\newblock In {\em Proceedings of the IEEE/CVF Conference on Computer Vision and Pattern Recognition}, 2025.

\bibitem{yang2023vid2seq}
Antoine Yang, Arsha Nagrani, Paul~Hongsuck Seo, Antoine Miech, Jordi Pont-Tuset, Ivan Laptev, Josef Sivic, and Cordelia Schmid.
\newblock Vid2seq: Large-scale pretraining of a visual language model for dense video captioning.
\newblock In {\em Proceedings of the IEEE/CVF Conference on Computer Vision and Pattern Recognition}, pages 10714--10726, 2023.

\bibitem{yang2023vid2seq_yt_temporal}
Antoine Yang, Arsha Nagrani, Paul~Hongsuck Seo, Antoine Miech, Jordi Pont-Tuset, Ivan Laptev, Josef Sivic, and Cordelia Schmid.
\newblock Vid2seq: Large-scale pretraining of a visual language model for dense video captioning.
\newblock In {\em Proceedings of the IEEE/CVF Conference on Computer Vision and Pattern Recognition}, pages 10714--10726, 2023.

\bibitem{yang2025egolife}
Jingkang Yang, Shuai Liu, Hongming Guo, Yuhao Dong, Xiamengwei Zhang, Sicheng Zhang, Pengyun Wang, Zitang Zhou, Binzhu Xie, Ziyue Wang, Bei Ouyang, Zhengyu Lin, Marco Cominelli, Zhongang Cai, Yuanhan Zhang, Peiyuan Zhang, Fangzhou Hong, Joerg Widmer, Francesco Gringoli, Lei Yang, Bo~Li, and Ziwei Liu.
\newblock Egolife: Towards egocentric life assistant.
\newblock In {\em Proceedings of the IEEE/CVF Conference on Computer Vision and Pattern Recognition}, 2025.

\bibitem{yu2025dapo}
Qiying Yu, Zheng Zhang, Ruofei Zhu, Yufeng Yuan, Xiaochen Zuo, Yu~Yue, Tiantian Fan, Gaohong Liu, Lingjun Liu, Xin Liu, et~al.
\newblock Dapo: An open-source llm reinforcement learning system at scale.
\newblock {\em arXiv preprint arXiv:2503.14476}, 2025.

\bibitem{yu2024rlhfv}
Tianyu Yu, Yuan Yao, Haoye Zhang, Taiwen He, Yifeng Han, Ganqu Cui, Jinyi Hu, Zhiyuan Liu, Hai-Tao Zheng, Maosong Sun, et~al.
\newblock Rlhf-v: Towards trustworthy mllms via behavior alignment from fine-grained correctional human feedback.
\newblock In {\em Proceedings of the IEEE/CVF Conference on Computer Vision and Pattern Recognition}, pages 13807--13816, 2024.

\bibitem{yuan2021tIoU}
Yitian Yuan, Xiaohan Lan, Xin Wang, Long Chen, Zhi Wang, and Wenwu Zhu.
\newblock A closer look at temporal sentence grounding in videos: Dataset and metric.
\newblock In {\em Proceedings of the 2nd international workshop on human-centric multimedia analysis}, pages 13--21, 2021.

\bibitem{zala2023hirest}
Abhay Zala, Jaemin Cho, Satwik Kottur, Xilun Chen, Barlas Oguz, Yashar Mehdad, and Mohit Bansal.
\newblock Hierarchical video-moment retrieval and step-captioning.
\newblock In {\em Proceedings of the IEEE/CVF Conference on Computer Vision and Pattern Recognition}, pages 23056--23065, 2023.

\bibitem{zeng2025timesuite}
Xiangyu Zeng, Kunchang Li, Chenting Wang, Xinhao Li, Tianxiang Jiang, Ziang Yan, Songze Li, Yansong Shi, Zhengrong Yue, Yi~Wang, Yali Wang, Yu~Qiao, and Limin Wang.
\newblock Timesuite: Improving {MLLM}s for long video understanding via grounded tuning.
\newblock In {\em The Thirteenth International Conference on Learning Representations}, 2025.

\bibitem{zhang2023tsgsurvey}
Hao Zhang, Aixin Sun, Wei Jing, and Joey~Tianyi Zhou.
\newblock Temporal sentence grounding in videos: A survey and future directions.
\newblock {\em IEEE Transactions on Pattern Analysis and Machine Intelligence}, 45(8):10443--10465, 2023.

\bibitem{zhang2021ms2dtan}
Songyang Zhang, Houwen Peng, Jianlong Fu, Yijuan Lu, and Jiebo Luo.
\newblock Multi-scale 2d temporal adjacency networks for moment localization with natural language.
\newblock {\em IEEE Transactions on Pattern Analysis and Machine Intelligence}, 2021.

\bibitem{zhang20192DTAN}
Songyang Zhang, Houwen Peng, Jianlong Fu, and Jiebo Luo.
\newblock Learning 2d temporal adjacent networks for moment localization with natural language.
\newblock In {\em Proceedings of the AAAI Conference on Artificial Intelligence}, 2020.

\bibitem{zhang2025tinyllavavideor1}
Xingjian Zhang, Siwei Wen, Wenjun Wu, and Lei Huang.
\newblock Tinyllava-video-r1: Towards smaller lmms for video reasoning.
\newblock {\em arXiv preprint arXiv:2504.09641}, 2025.

\bibitem{zhao2025videoexpert}
Henghao Zhao, Ge-Peng Ji, Rui Yan, Huan Xiong, and Zechao Li.
\newblock Videoexpert: Augmented llm for temporal-sensitive video understanding.
\newblock {\em arXiv preprint arXiv:2504.07519}, 2025.

\bibitem{ssrn}
Jiahao Zhu, Daizong Liu, Pan Zhou, Xing Di, Yu~Cheng, Song Yang, Wenzheng Xu, Zichuan Xu, Yao Wan, Lichao Sun, and Zeyu Xiong.
\newblock Rethinking the video sampling and reasoning strategies for temporal sentence grounding.
\newblock In {\em Findings of the Association for Computational Linguistics: EMNLP 2022}, 2022.

\end{thebibliography}
}
\appendix

\newpage

\startcontents
\printcontents{}{1}{}

\section{Limitations}
\label{sec:app_discuss}
% \noindent\textbf{. }
Despite achieving notable improvements on the TVG task, our approach still has several limitations. 
First, Time-R1 suffers from slower training and inference speeds, primarily due to its large model size and reliance on autoregressive text generation. 
Second, to manage GPU memory consumption, we use a relatively low frame sampling rate, which may result in the loss of fine-grained motion information across frames. 
Finally, Time-R1 currently cannot handle ultra-long videos, limiting its applicability in scenarios such as full-length movie understanding.

\section{Implementation Details}
\label{sec:imple_details}

\noindent\textbf{Details of Time-R1 framework.}
Inspired by DAPO~\cite{yu2025dapo}, we adopt its token-level loss for training, rather than the sample-level loss used in GRPO.
Apart from minor changes to the loss, all settings are identical to GRPO.
Besides, we find that other techniques introduced in DAPO do not benefit the TVG task, thus aborting other techniques.
We full-finetune the LLM parameters at every step, thus $\frac{\pi_\theta(o_i)}{\pi_{\theta_{\mathrm{old}}}(o_i)}=1$.
The sample number $G$ is set to 8. The coefficient $\beta$ is set to 0.04.

\noindent\textbf{Details of TimeRFT training.}
For RFT data filtering, we use a Gaussian distribution with a fixed variance of 0.2 while varying the mean to control sample selection.
In our cold start phase, we construct 150 samples from our training data sources (e.g., YT-Temporal~\cite{yang2023vid2seq_yt_temporal}) to fine-tune the LLM using LoRA~\cite{hu2022lora}, with a LoRA rank of 64 and a LoRA alpha of 128.
All of our results are reported based on the final training epoch.
For RL, we use a learning rate of 1e-6 with the AdamW optimizer with $\beta_1$=0.9, $\beta_2$ = 0.999, and a linear scheduler to decay the learning rate from 1e-6 to 0.
We use a batch size of 8 with gradient accumulation set to 2.
It requires 15 hours of training on 8 A100 GPUs.

\noindent\textbf{Details of our evaluation prompts.}
As shown in~\Cref{fig:prompt_illustration}, for temporal video grounding, the prompts used for training and testing are designed to encourage the model to reason before responding, following a template-based answer format. For VideoQA, we have two versions of prompts: one with CoT and one without CoT.

\noindent\textbf{Details of TVG baseline methods and implementations.}
We evaluate the baselines on TVGBench using their original best-performing setting, focusing primarily on video input and prompt design.
\begin{itemize}[nosep,wide,labelindent=0pt,labelwidth=*,align=left]
    \item TimeChat~\cite{ren2024timechat} is built upon the InstructBLIP~\cite{instructblip} architecture and introduces a video Q-former to encode video tokens. It operates at a resolution of 224 and samples 96 frames.
    \item TRACE~\cite{guo2024trace} treats each combination of timestamp, saliency score, and caption as a discrete event and enables the LVLM to autoregressively generate event sequences. It operates at a higher resolution of 336 and samples 128 frames.
    \item TimeSuite~\cite{zeng2025timesuite} introduces a token shuffling strategy to compress long video token sequences and incorporates positional encoding to enhance visual understanding. It adopts a resolution of 224 and samples 128 frames.
    \item VideoChat-Flash~\cite{li2024videochatflash} proposes a progressive visual token dropping mechanism within intermediate LLM layers to compress video inputs and extend the effective context length. It uses a resolution of 448 and samples video at 1 fps, with a maximum of 512 frames.
    \item Gemini-2.5-Pro~\cite{Gemini2.5}: Gemini-2.5-Pro is a state-of-the-art video understanding model capable of reasoning over videos exceeding one hour in length. It supports video question answering and temporal localization tasks.
\end{itemize}

\noindent\textbf{Details of our implemented SFT baselines.}
We implemented two versions of SFT fine-tuning: one is full-parameter fine-tuning of the LLM (SFT), and the other is LoRA-based fine-tuning of the LLM (SFT-LoRA). 
For SFT-LoRA, the LoRA rank is set to 64, and the LoRA alpha is set to 128. Both configurations use the following settings: a learning rate of 2e-5, the AdamW optimizer with $\beta_1$=0.9, $\beta_2$ = 0.999, a weight decay of 0, the batch size of 8, and accumulation steps of 2.
We fine-tune for 5 epochs on our 2.5K data, and use a linear scheduler to gradually decay the learning rate to 0.

\section{Additional Analyses}

\begin{table*}[t]
\centering
\caption{Comparison of different approaches on TVGBench for all types. We use mIoU as metric.
}
\setlength{\belowcaptionskip}{3pt}%
\scalebox{0.8}{
\begin{tabular}{l|cccccccccccc}
\toprule
Method & EC & ES & HAC & HAP & HAS & HP & OA & OC & OEC & OES & OT \\
\midrule
TimeChat~\cite{ren2024timechat} & 22.3 & 32.8 & 16.6 & 9.8 & 14.6 & 35.1 & 15.0 & 9.2 & 2.4 & 18.0 & 10.2 \\
TimeSuite~\cite{zeng2025timesuite} & 27.3 & 39.6 & 14.2 & 12.8 & 24.9 & 39.6 & 14.6 & 13.9 & 6.7 & 32.6 & 14.3 \\
TRACE~\cite{guo2024trace} & 57.1 & 66.8 & 25.9 & 17.5 & 26.5 & 45.1 & 17.8 & 22.1 & 12.5 & 36.8 & 24.9 \\
VideoChat-Flash~\cite{li2024videochatflash} & 38.3 & 47.2 & 12.9 & 13.9 & 27.1 & 39.4 & 14.9 & 12.7 & 6.5 & 24.3 & 12.9 \\
Gemini-2.5-Pro~\cite{Gemini2.5} & 46.7 & 45.3 & 21.1 & 27.6 & 30.9 & 39.9 & 23.0 & 31.1 & 14.1 & 35.9 & 17.8 \\
\midrule
Time-R1 (ours) & 49.3 & 65.3 & 28.3 & 24.3 & 39.3 & 56.2 & 26.3 & 21.8 & 9.0 & 32.7 & 21.8 \\  

\bottomrule
\end{tabular}}
\label{tab:sup_alltype_TVGBench}
\end{table*}

\noindent\textbf{In-depth comparisons of different approaches on TVGBench by semantic type.}
~\Cref{tab:sup_alltype_TVGBench} provides a detailed performance comparison of various methods on the TVGBench across different semantic categories. 
Specifically, the abbreviations represent: EC (Environment Change), ES (Environment State), HAC (Human Action – Complex), HAP (Human Action – Procedural), HAS (Human Action – Simple), HP (Human Pose), OA (Object Attribute), OC (Object Counting), OEC (Object Existence – Complex), OES (Object Existence – Simple), and OT (Object Transition). 
Detailed definition and construction process can be found in~\Cref{fig:prompt_illustration_tvgbench_semantics}.

Time-R1 demonstrates strong competitiveness across multiple semantic categories. 
First, particularly in the four tasks of HAC, HAS, HP, and OA, Time-R1 achieved the highest scores among all compared methods, showcasing its excellent ability in understanding the details of human actions and identifying object features. 
For example, Time-R1 achieves an mIoU of 56.2 on HP, which is 11.1 points higher than the second-best method, TRACE, with an mIoU of 45.1. 
On HAS, Time-R1 reaches 39.3, outperforming Gemini-2.5-Pro’s 30.9 by 8.4 points.
Second, in the three tasks of ES, EC, and OT, Time-R1 demonstrates strong performance comparable to the top model TRACE, with its performance being very close or immediately following. 
In the HAP task, Time-R1 also performs excellently, with its performance being in the same tier as Gemini-2.5-Pro.
Lastly, all models still show a noticeable gap compared to Gemini in understanding complex instructions, such as in HAP, OC, and OEC.
For example, in HAP, which involves procedural activity localization, Gemini achieves 27.6, while our model ranks second with a score of 24.3. In object counting, Gemini attains 31.1, substantially outperforming our model's 21.8.
In summary, Time-R1 performs well on both non-human simple instructions and human-related instructions, but there is still room for improvement in complex instruction grounding and object-related grounding.

\noindent\textbf{Comparison of speed and accuracy between inference library transformers and vLLM.}
We observe that the inference speed of the implementation in the transformers~\cite{transformers_lib} library is very slow. 
To address this, we implemented an accelerated inference version using vLLM~\cite{kwon2023vllm} for all related 7 downstream benchmarks. 
For example, on TVGBench, the vLLM-based implementation requires only 502 seconds to infer 800 samples using 8 GPUs, whereas the transformers library implementation takes 2520 seconds. 
This achieves an overall speedup of 5$\times$.

\section{Ablation Studies}
\label{sec: abl_study}

\begin{table*}[t]
\setlength{\tabcolsep}{5pt}
\begin{floatrow}
\capbtabbox{
\scalebox{0.7}{
\begin{tabular}{c|cccc}
\toprule
 Method & \multicolumn{1}{c}{R1@0.3} & \multicolumn{1}{c}{R1@0.5} & \multicolumn{1}{c}{R1@0.7} & \multicolumn{1}{c}{mIoU}\\
\midrule
random         & 39.4 & 26.5 & 16.4 & 27.4 \\
gaussian (0.3) & 41.6 & 28.5 & 15.6 & 28.6 \\
gaussian (0.5) & 40.6 & 28.2 & 16.0 & 28.3 \\
gaussian (0.7) & 37.2 & 26.9 & 15.5 & 26.5 \\
uniform        & 40.4 & 28.5 & 15.9 & 28.3 \\

\bottomrule
\end{tabular}
}
}{
 \caption{Ablation of data filtering strategies.}
\label{tab:abl_RFT_filtering_strategy}
}

\capbtabbox{
\scalebox{0.81666}{
\begin{tabular}{cc|cccc}
\toprule
 KL & CoT & \multicolumn{1}{c}{R1@0.3} & \multicolumn{1}{c}{R1@0.5} & \multicolumn{1}{c}{R1@0.7} & \multicolumn{1}{c}{mIoU}\\
\midrule
\ding{55} & \ding{55} & 40.4 & 29.1 & 14.9 & 28.1 \\
\ding{51} & \ding{55} & 40.8 & 27.4 & 15.0 & 27.7 \\
\ding{55} & \ding{51} & 42.9 & 29.5 & 15.0 & 29.1 \\
\ding{51} & \ding{51} & 41.6 & 28.5 & 15.6 & 28.6 \\

\bottomrule
\end{tabular}
}
}{
 \caption{Ablation of KL and CoT in GRPO.}
 \label{tab:abl_kl_cot}
 \small
}
\end{floatrow}
\end{table*}

\begin{table*}[t]
\centering
\caption{Comparison of the token-level loss design used by DAPO~\cite{yu2025dapo} and the sample-level loss design used by GRPO~\cite{grpo}.
}
\setlength{\belowcaptionskip}{3pt}%
\scalebox{0.73}{
\begin{tabular}{l|ccccc@{}|ccccc@{}|cccc}
\toprule
\multirow{2}{*}{Loss} & \multicolumn{4}{c}{Charades-STA} && \multicolumn{4}{c}{ActivityNet} && \multicolumn{4}{c}{TVGBench}\\
 & {\fontsize{8.4}{10}\selectfont R1@0.3} & {\fontsize{8.4}{10}\selectfont R1@0.5} & {\fontsize{8.4}{10}\selectfont R1@0.7} & {\fontsize{8.4}{10}\selectfont mIoU} && {\fontsize{8.4}{10}\selectfont R1@0.3} & {\fontsize{8.4}{10}\selectfont R1@0.5} & {\fontsize{8.4}{10}\selectfont R1@0.7} & {\fontsize{8.4}{10}\selectfont mIoU} && {\fontsize{8.4}{10}\selectfont R1@0.3} & {\fontsize{8.4}{10}\selectfont R1@0.5} & {\fontsize{8.4}{10}\selectfont R1@0.7} & {\fontsize{8.4}{10}\selectfont mIoU} \\
\midrule
GRPO & 
76.7 & 59.8 & 34.4 & 57.0 && 55.9 & 37.1 & 20.3 & 37.8 && 40.8 & 28.0 & 16.5 & 28.4 \\  
DAPO & 
77.4 & 60.0 & 34.1 & 57.2 && 56.2 & 37.4 & 20.4 & 38.0 && 41.6 & 28.5 & 15.6 & 28.6 \\  

\bottomrule
\end{tabular}}
\label{tab:sup_dapo_grpo}
\end{table*}

\begin{table*}[t]
\centering
\caption{Performance comparison of different model sizes.
}
\setlength{\belowcaptionskip}{3pt}%
\scalebox{0.73}{
\begin{tabular}{l|ccccc@{}|ccccc@{}|cccc}
\toprule
\multirow{2}{*}{Method} & \multicolumn{4}{c}{Charades-STA} && \multicolumn{4}{c}{ActivityNet} && \multicolumn{4}{c}{TVGBench}\\
 & {\fontsize{8.4}{10}\selectfont R1@0.3} & {\fontsize{8.4}{10}\selectfont R1@0.5} & {\fontsize{8.4}{10}\selectfont R1@0.7} & {\fontsize{8.4}{10}\selectfont mIoU} && {\fontsize{8.4}{10}\selectfont R1@0.3} & {\fontsize{8.4}{10}\selectfont R1@0.5} & {\fontsize{8.4}{10}\selectfont R1@0.7} & {\fontsize{8.4}{10}\selectfont mIoU} && {\fontsize{8.4}{10}\selectfont R1@0.3} & {\fontsize{8.4}{10}\selectfont R1@0.5} & {\fontsize{8.4}{10}\selectfont R1@0.7} & {\fontsize{8.4}{10}\selectfont mIoU} \\
\midrule
Time-R1-3B & 
74.6 & 53.1 & 26.0 & 51.2 && 40.0 & 21.0 & 8.7& 23.2 && 33.5 & 21.0 & 10.5& 21.7 \\  
Time-R1-3B$^*$ & 
78.7 & 64.1 & 36.9& 59.9 && 66.8 & 46.8 & 24.7& 46.1 && - & - & -& - \\  
\midrule
Time-R1-7B & 
78.1 & 60.8 & 35.5& 58.1 && 58.1 & 39.0 & 21.4& 40.5 && 41.8 & 29.4 & 16.4 & 29.2 \\
Time-R1-7B$^*$ & 
82.8 & 72.2 & 50.1 & 60.9 && 73.3 & 55.6 & 34.0 & 52.1 && - & - & -& - \\

\bottomrule
\end{tabular}}
\label{tab:sup_ModelSize}
\end{table*}

\noindent\textbf{Ablation of different RFT data filtering strategies.}
As shown in~\Cref{tab:abl_RFT_filtering_strategy}, different data filtering strategy in the initial round affects the model's performance. 
First, appropriate Gaussian filtering outperforms both uniform and random filtering methods.
Among the Gaussian filtering settings, a standard deviation of 0.3 yields the best results, followed by 0.5 and then 0.7.
These findings suggest that incorporating moderately challenging samples during RFT helps improve the model’s generalization capability more effectively than using either overly easy or extremely difficult examples.

\noindent\textbf{Ablation of KL and CoT during GRPO training.}
As shown in~\Cref{tab:abl_kl_cot}, incorporating CoT reasoning during training leads to improved performance compared to the No-CoT setting, suggesting that CoT enhances the model's temporal video grounding capabilities. 
When KL divergence is omitted (No-KL), performance slightly decreases under the No-CoT setting but unexpectedly improves when CoT is present. 
However, we find that in the No-KL+CoT setting, the model often fails to produce a thinking process, directly jumping to answers.
In contrast, using KL divergence helps maintain more logical reasoning that is easier to follow.
To balance performance and interpretability, we adopt a training setup that includes both KL and CoT.

\begin{figure*}[t]
\begin{floatrow}
    \centering
    \capbfigbox{
        \begin{minipage}{0.45\textwidth}
            \centering
            \includegraphics[width=\linewidth]{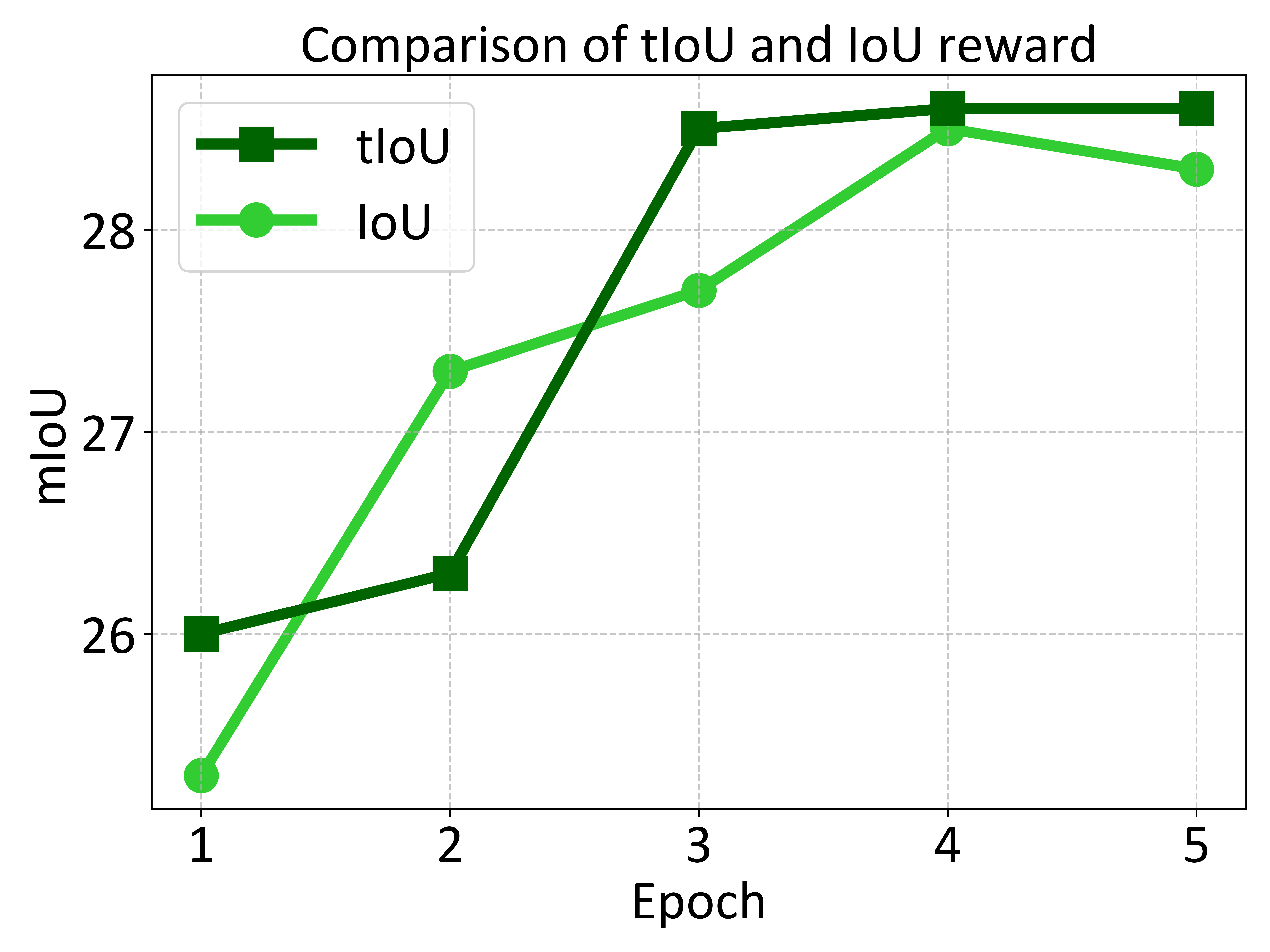}
        \end{minipage}
    }{
     \caption{Performance comparison of tIoU and IoU in multi-epoch training.}
     \label{fig:tIoU}
     \small
    }
    \hfill
    \capbfigbox{
        \begin{minipage}{0.45\textwidth}
            \centering
            \includegraphics[width=\linewidth]{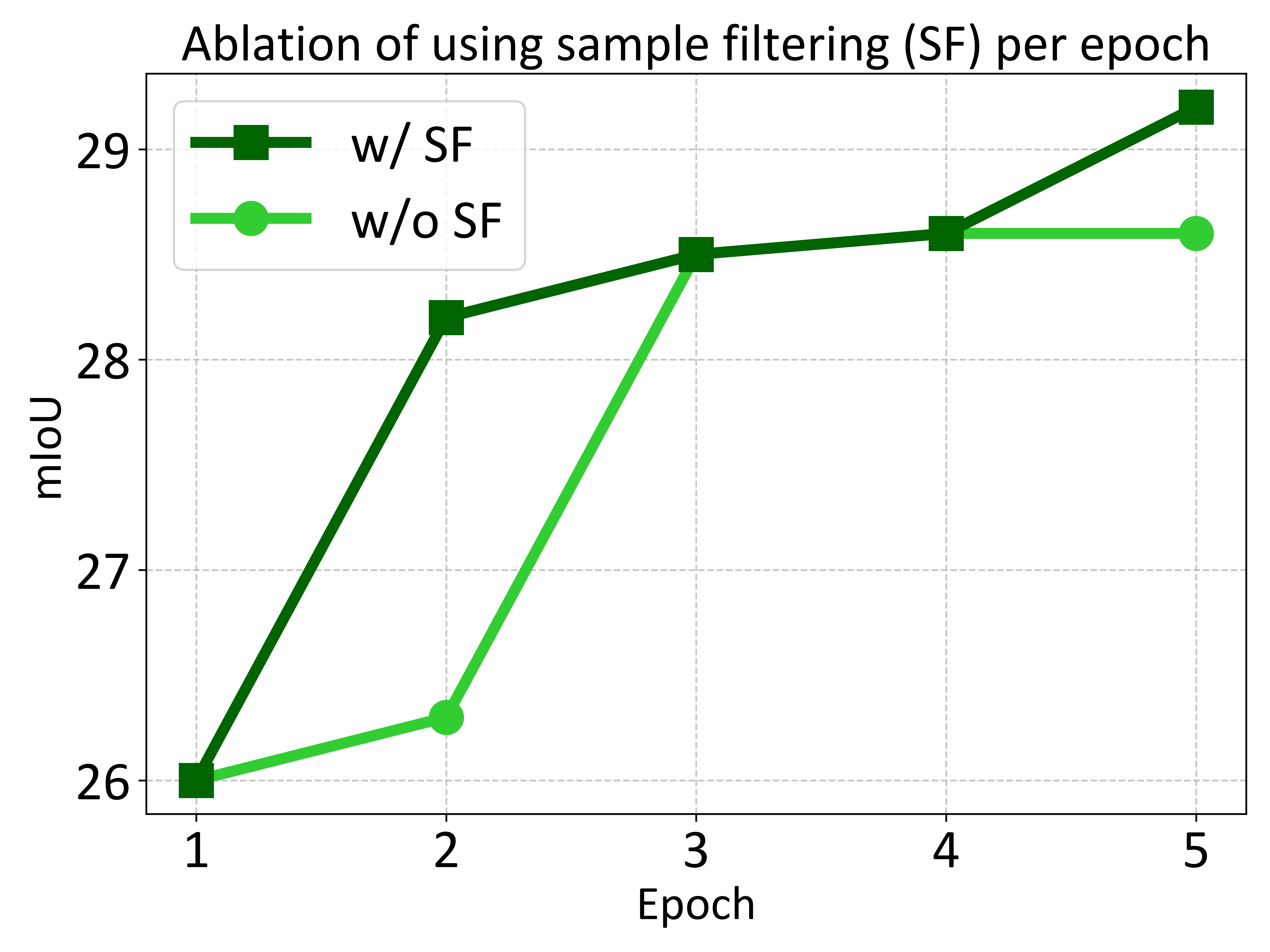}
        \end{minipage}
    }{
     \caption{Ablation of sample filtering in multi-epoch training.}
     \label{fig:data_filtering}
     \small
    }
\end{floatrow}    
\end{figure*}

\noindent\textbf{Comparison of tIoU and IoU during multi-epoch training.}
As shown in~\Cref{fig:tIoU}, tIoU consistently outperforms standard IoU during both the early and late stages of training over the first 5 epochs. Notably, while tIoU steadily improves as training progresses, IoU shows a decline in performance by the fifth epoch. This highlights the advantage of using tIoU as a more stable and reliable reward for temporal video grounding.

\noindent\textbf{Ablation of sample filtering in multi-epoch training.}
As shown in~\Cref{fig:data_filtering}, applying sample filtering (SF) to remove simpler training samples yields consistent performance improvements across epochs. This suggests that easy samples with high IoU may introduce noise or reduce the effectiveness of learning, and filtering them helps focus the model on more informative and challenging instances.

\noindent\textbf{Ablation of DAPO \& GRPO.}
The sample-level loss used by GRPO computes the loss by averaging over each individual sample.
This approach leads to unequal loss contributions for tokens when dealing with CoTs of varying lengths. 
DAPO addresses this issue by employing a token-level loss. 
The underlying principle is that the token-level loss can effectively guide the model in the process of CoT generation, allowing it to learn useful patterns from CoTs of different lengths sampled during training.
In~\Cref{tab:sup_dapo_grpo}, we compare these two loss designs.
We empirically find that DAPO outperforms GRPO on the majority of metrics, thus, we adopt DAPO's loss design.

\noindent\textbf{Different Model Size.}
~\Cref{tab:sup_ModelSize} presents a performance comparison of different model sizes. 
These results indicate that larger models achieve better zero-shot performance and continue to outperform smaller models after fine-tuning.  
These findings support the notion that scaling up model capacity enhances generalization and leads to superior results on the TVG tasks.

\section{Qualitative Result}
\label{sec:quanli_result}

\begin{figure}[t]
    \centering
    \includegraphics[width=\linewidth]{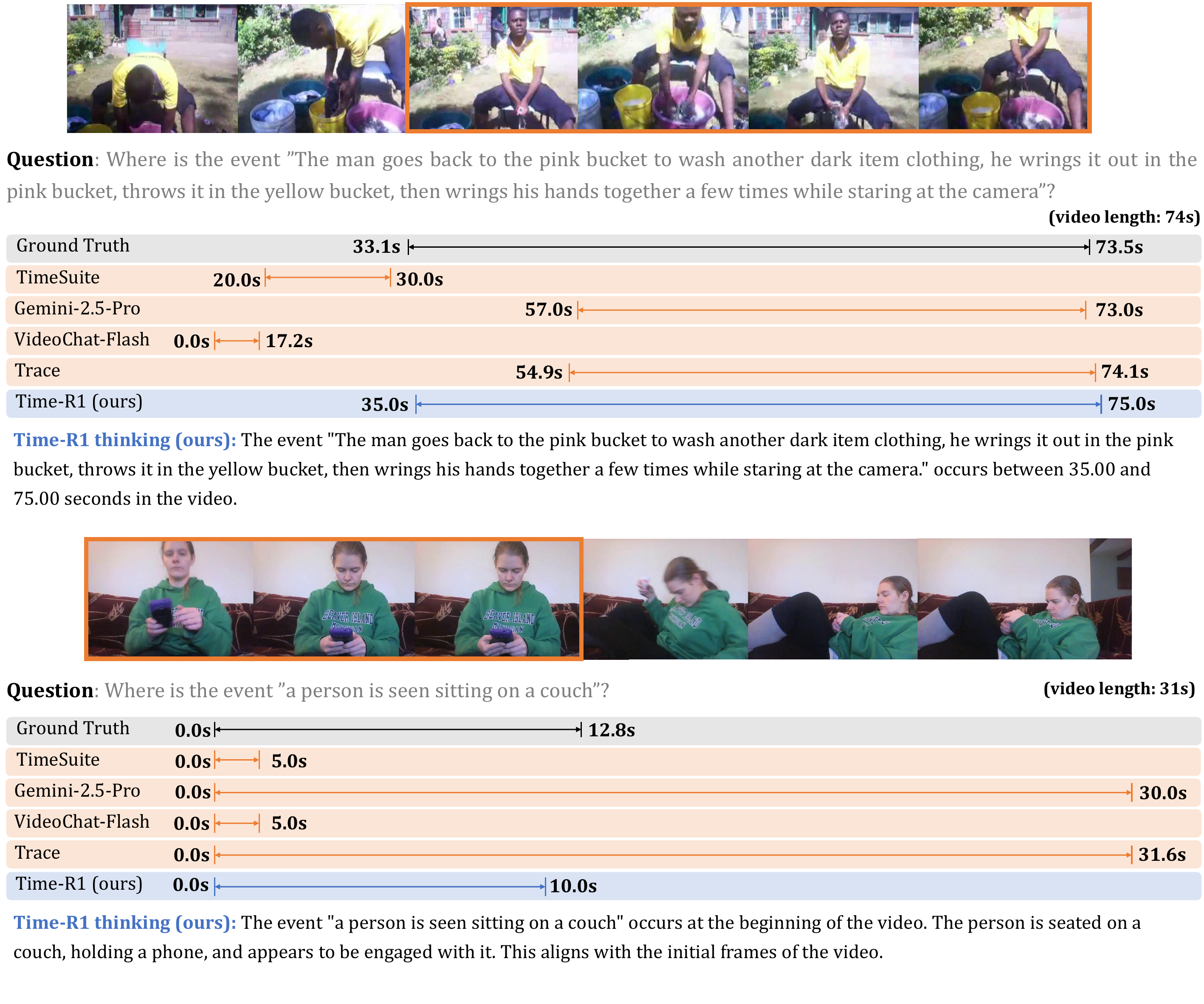}
    \caption{Temporal video grounding cases from Charades and ActivityNet benchmark. {\color{blue}Time-R1} outperforms {\color{orange}other SoTA} models.}
    \label{fig:case_tvgbench}
\end{figure}

\noindent\textbf{Case study of temporal video grounding on Charades and ActivityNet.}
As shown in~\Cref{fig:case_tvgbench}, in the example above, given a relatively complex language instruction, Time-R1 demonstrates more accurate localization than all baselines, successfully capturing the initial event “goes back to the pink bucket” within the timestamp, while other SoTA models like Gemini-2.5-Pro fail.
In the example below, the model accurately localizes the event, excluding “a person is lying on the couch,” and correctly distinguishes between sitting and lying, unlike other models, which either localize only a small segment (TimeSuite and VideoChat-Flash) or the entire segment (TRACE and Gemini-2.5-Pro).

\begin{figure}[h]
    \centering
    \includegraphics[width=\linewidth]{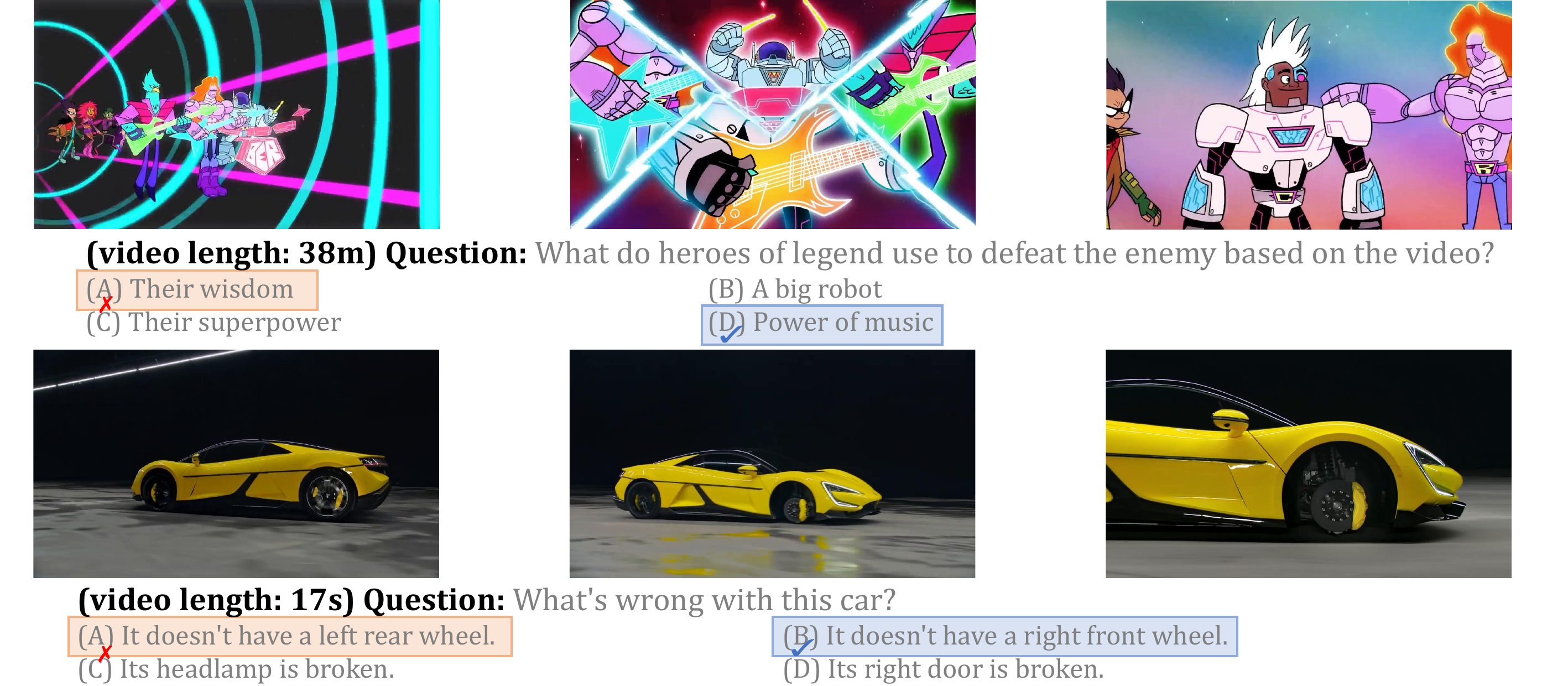}
    \caption{Case study on VideoMME (w/o CoT), demonstrating that {\color{blue}Time-R1} achieves better performance than the {\color{orange}base} model.}
    \label{fig:case_VideoMME}
\end{figure}

\begin{figure}[t]
    \centering
    \includegraphics[width=\linewidth]{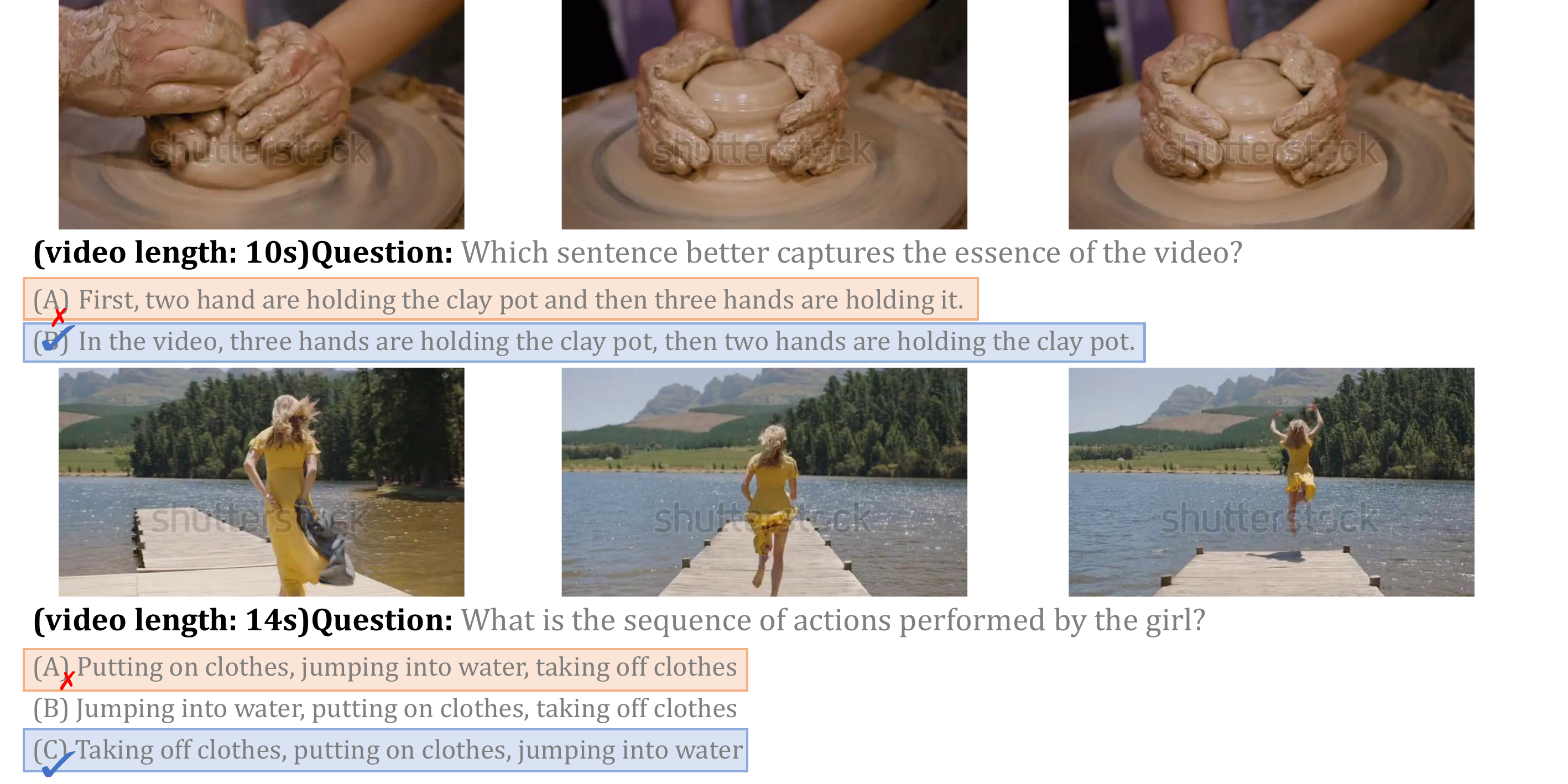}
    \caption{Case study on TempCompass (w/o CoT), demonstrating that {\color{blue}Time-R1} achieves better performance than the {\color{orange}base} model.}
    \label{fig:case_tempcompass}
\end{figure}

\begin{figure}[t]
    \centering
    \includegraphics[width=\linewidth]{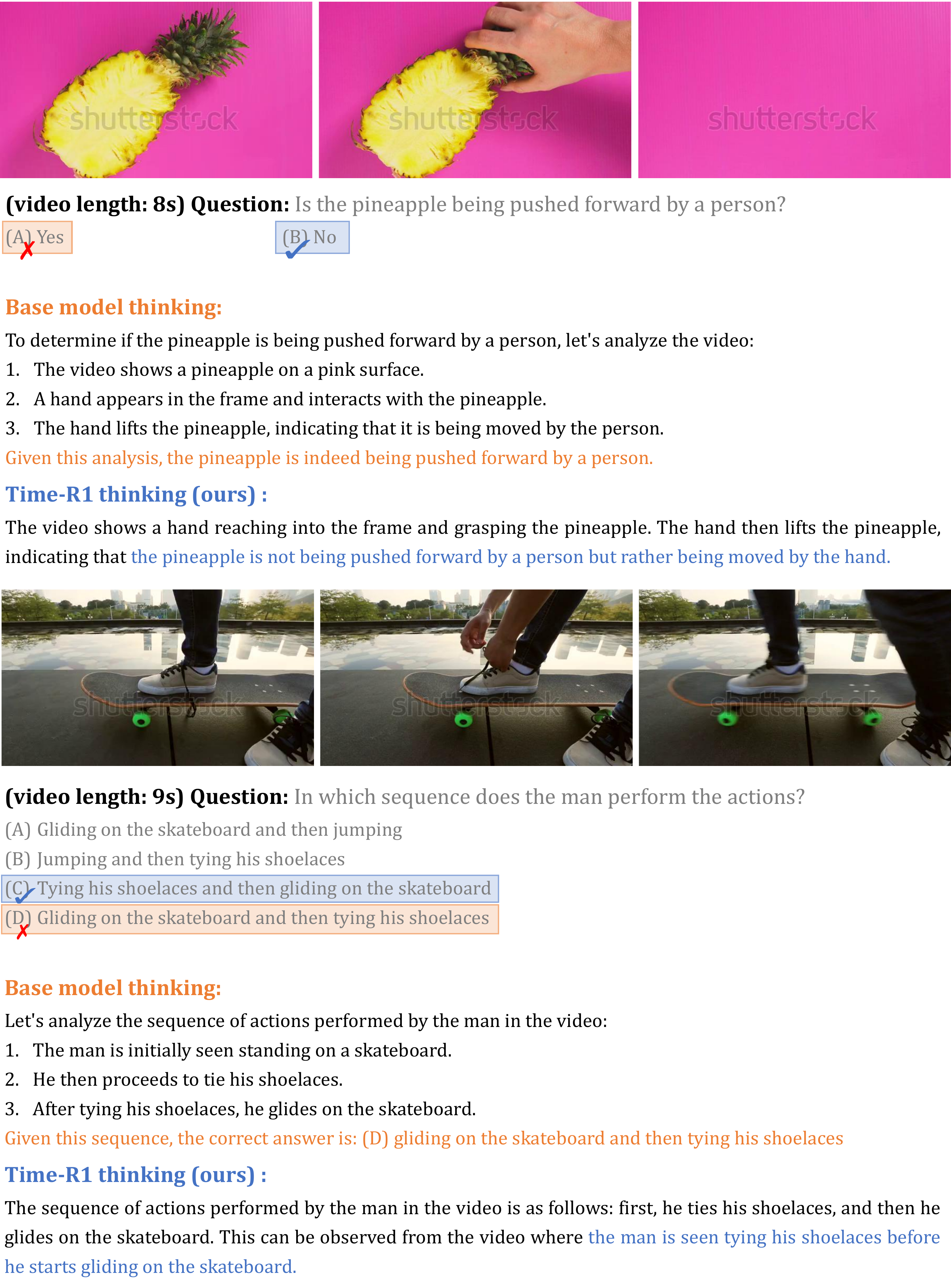}
    \caption{Case study on TempCompass (w/ CoT), demonstrating that {\color{blue}Time-R1} achieves better performance than the {\color{orange}base} model.}
    \label{fig:case_tempcompass_CoT}
\end{figure}

\noindent\textbf{Case study of short video QA on VideoMME and TempCompass.}
As shown in~\Cref{fig:case_VideoMME,fig:case_tempcompass}, Time-R1 demonstrates improved performance over the base model in tasks requiring positional judgment, scene storyline reasoning, and visual reasoning.
For example,  in~\Cref{fig:case_VideoMME}, Time-R1 correctly identifies that a car in the video is missing its right-front wheel, a detail that the base model fails to recognize.
This reflects that Time-R1 likely possesses stronger video localization capabilities, which in turn enhance its visual reasoning ability.
In~\Cref{fig:case_tempcompass_CoT}, we output a CoT when answering the QA task, providing some interpretability. 
This example shows that Time-R1’s reasoning process is more concise, whereas the base model often reasons correctly but arrives at the wrong answer.
This suggests that Time-R1’s reasoning may be more effective in guiding the final answer, possibly benefiting from the outcome-driven RL of GRPO.

\begin{figure}[t]
    \centering
    \includegraphics[width=\linewidth]{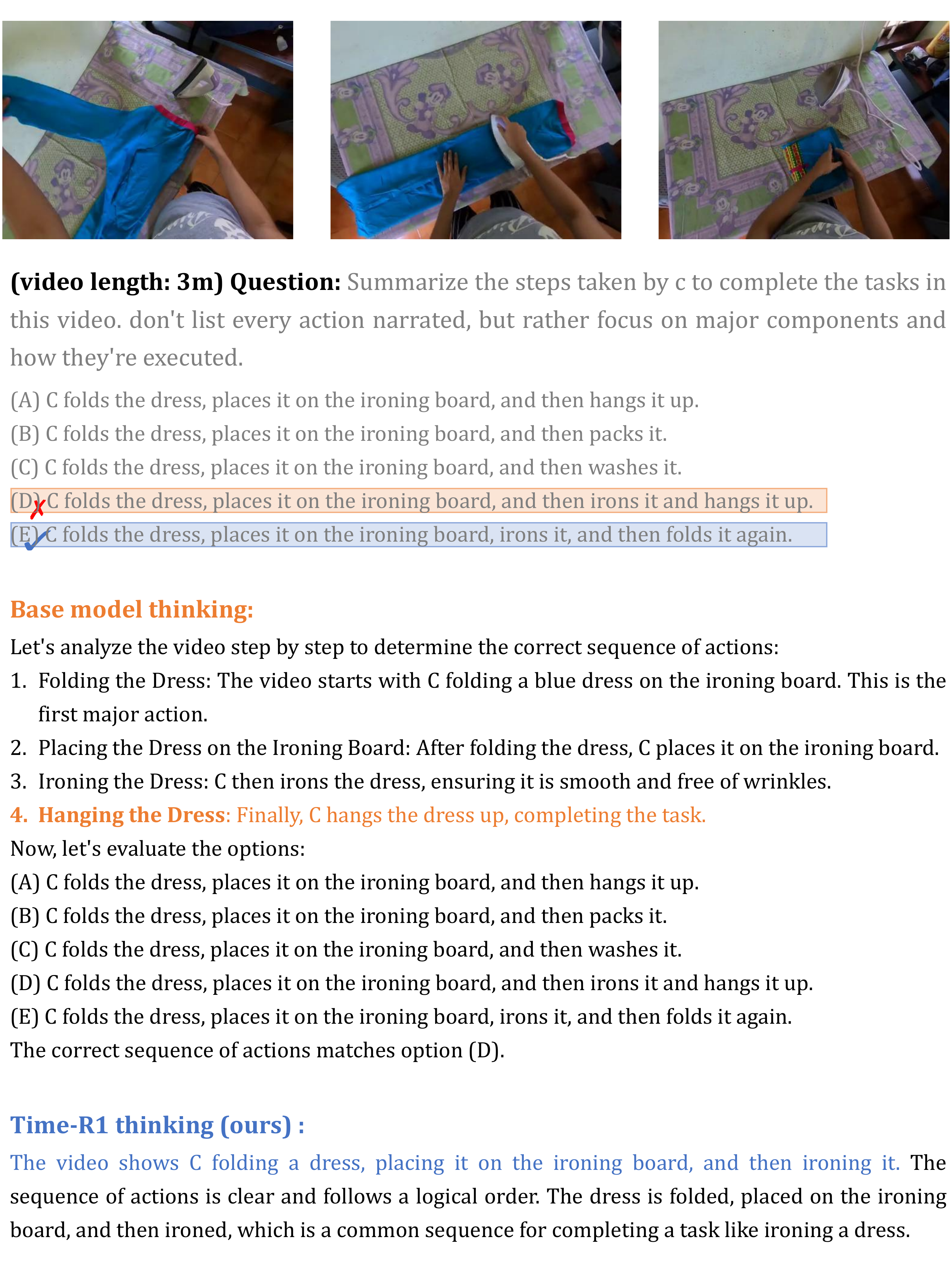}
    \caption{Case study on EgoSchema (w/ CoT), demonstrating that {\color{blue}Time-R1} achieves better performance than the {\color{orange}base} model.}
    \label{fig:case_egoschema1}
\end{figure}

\begin{figure}[t]
    \centering
    \includegraphics[width=\linewidth]{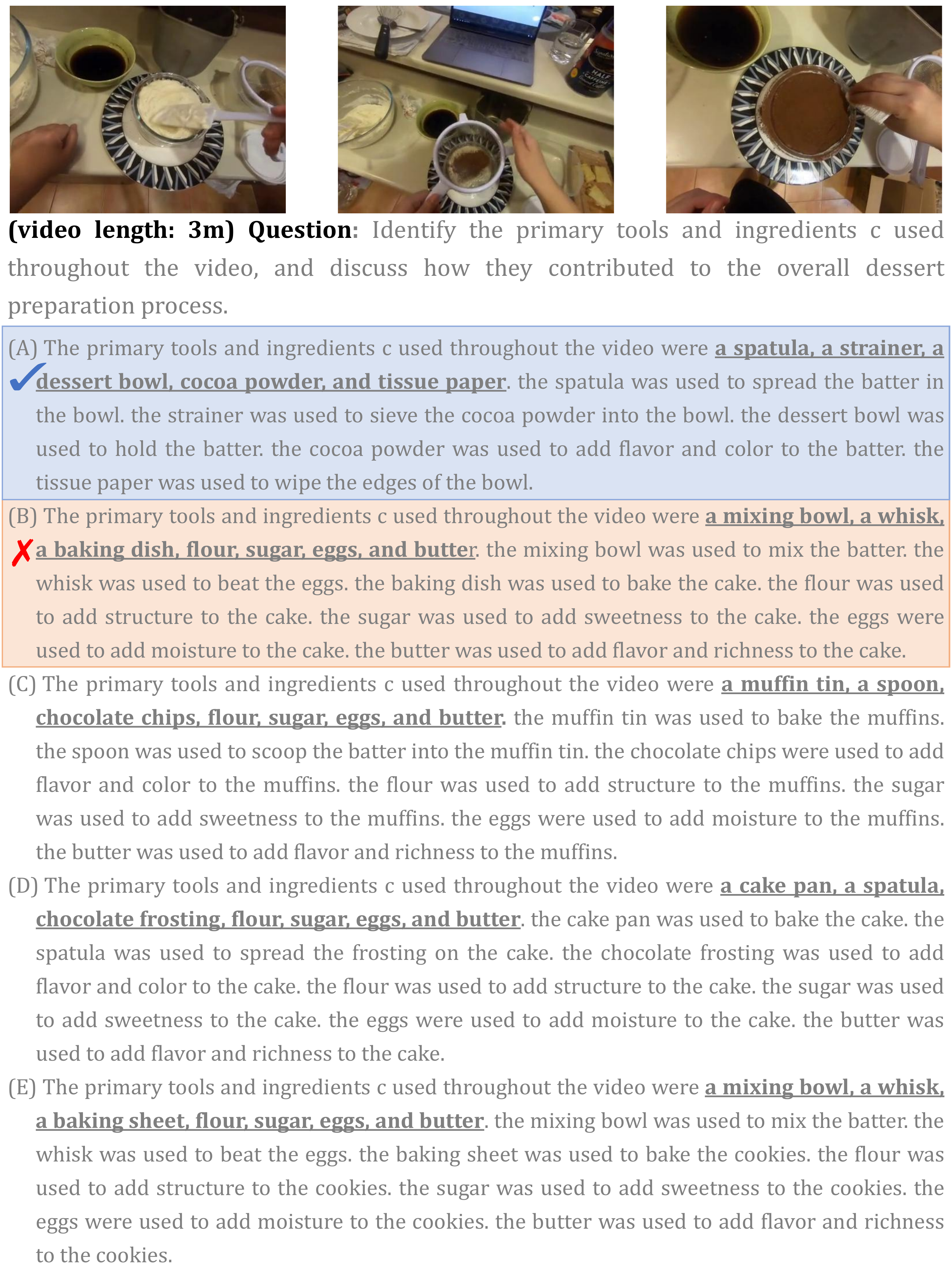}
    \caption{Case study on EgoSchema (w/o CoT), demonstrating that {\color{blue}Time-R1} achieves  better performance than the {\color{orange}base} model.}
    \label{fig:case_egoschema2}
\end{figure}

\noindent\textbf{Case study of long video QA on EgoSchema and VideoMME.}
~\Cref{fig:case_egoschema1} presents a long egocentric video QA example focused on summarizing task steps. 
In the "Hanging the Dress" case, the base model fails to identify all key steps, while our Time-R1 model correctly selects the answer by generating a more accurate chain-of-thought (CoT).
In~\Cref{fig:case_egoschema2},  the task involves identifying the primary tools and ingredients used in the video and understanding their respective contributions.
An additional example in~\Cref{fig:case_VideoMME} involves animated scene reasoning, where Time-R1 correctly infers, based on elements like the presence of a guitar, that the heroes defeated the enemies using the power of music, whereas the base model misinterprets the cause.
Across these complex reasoning tasks involving long videos, Time-R1 consistently demonstrates superior performance compared to the base model.

\begin{figure}[t]
    \centering
    \includegraphics[width=\linewidth]{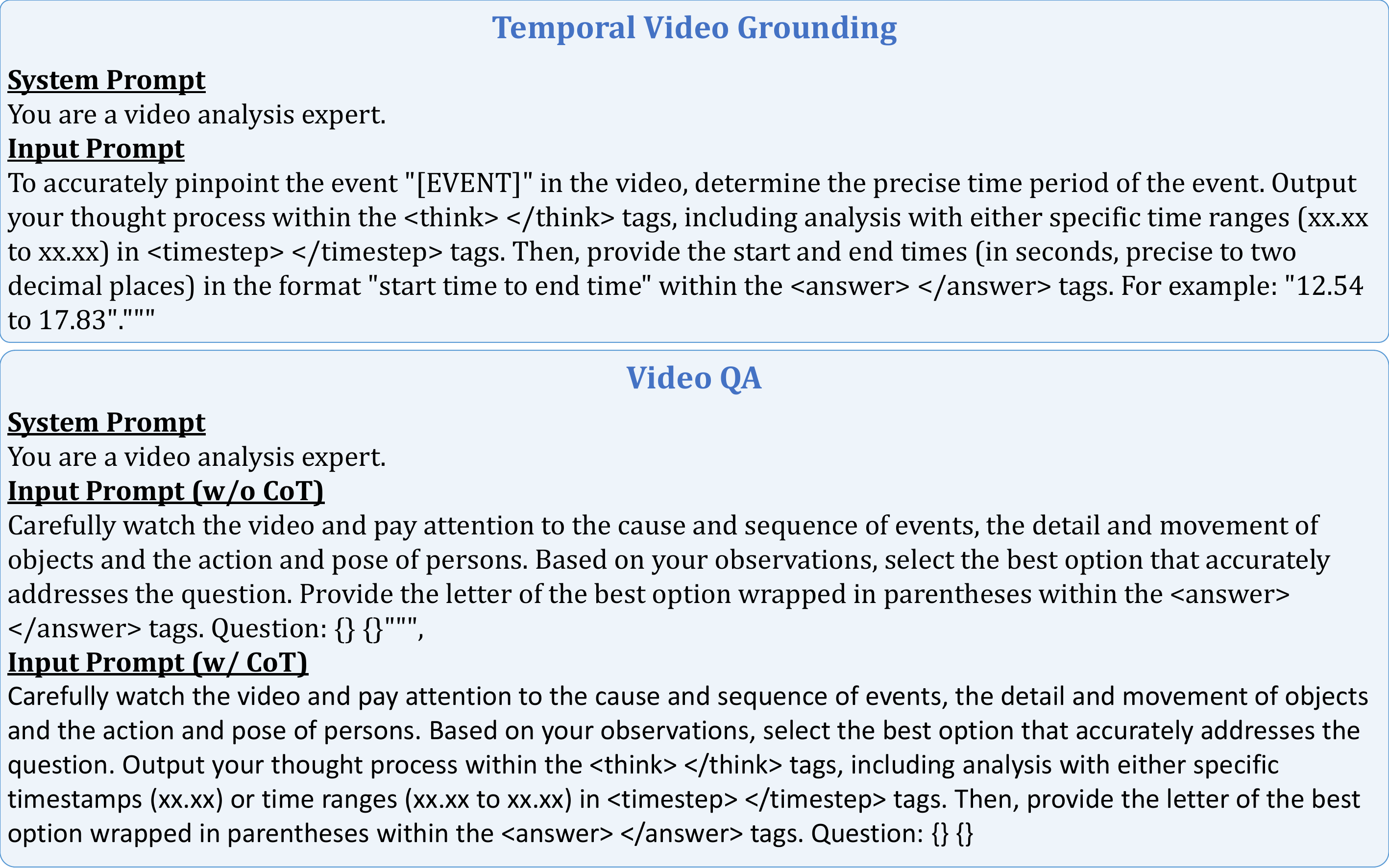}
    \caption{Illustration of prompts at both training and inference time.}
    \label{fig:prompt_illustration}
\end{figure}

\noindent\textbf{Illustration of our prompt at training and inference time.}
~\Cref{fig:prompt_illustration} presents the prompts used for the temporal video grounding and video QA tasks at both training and inference time.

\begin{figure}[t]
    \centering
    \includegraphics[width=\linewidth]{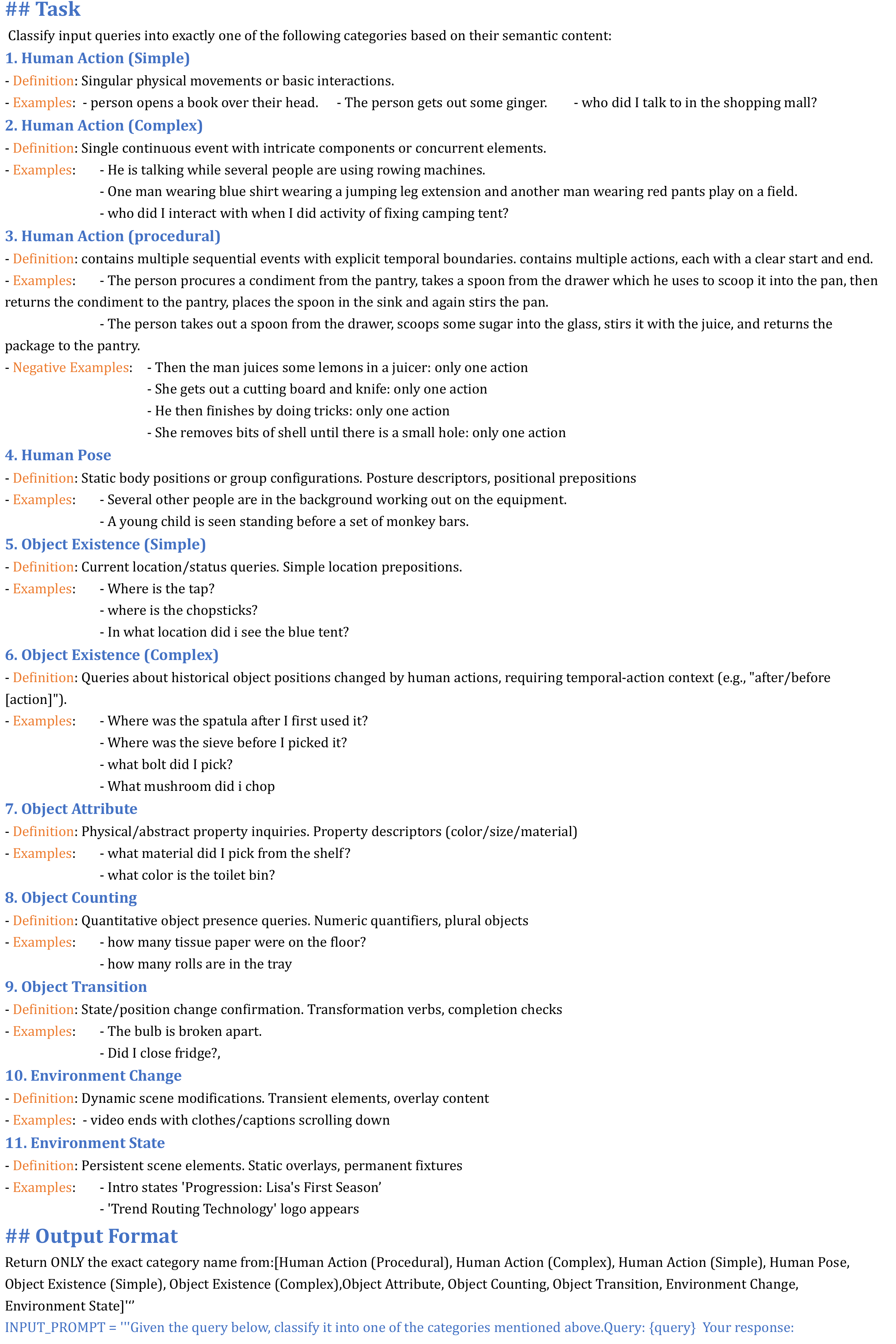}
    \caption{Prompts for LLM used to annotate the semantics of each query on TVGBench.}
    \label{fig:prompt_illustration_tvgbench_semantics}
\end{figure}

\noindent\textbf{Illustration of our prompt to annotate query semantics on TVGBench.}
~\Cref{fig:prompt_illustration_tvgbench_semantics} presents the prompt used to annotate query semantics in the TVGBench dataset.
The prompt is designed to guide the DeepSeek LLM in classifying each input query into one of 11 predefined semantic categories.
To improve annotation quality, we refer to the construction of existing benchmarks and carefully select both positive and negative examples for each semantic type.
These examples are chosen to include queries that are either easy or difficult for the model to answer, helping to refine the model’s understanding and improve labeling accuracy.

\clearpage
\newpage

\clearpage
\newpage

\end{document}